\documentclass{defaltmlr}

\usepackage{amsmath,amsfonts,bm}

\def\secref#1{section~\ref{#1}}

\def\eqref#1{equation~\ref{#1}}

\def\1{\bm{1}}

\DeclareMathAlphabet{\mathsfit}{\encodingdefault}{\sfdefault}{m}{sl}
\SetMathAlphabet{\mathsfit}{bold}{\encodingdefault}{\sfdefault}{bx}{n}

\usepackage{url}
\usepackage[table]{xcolor}
\usepackage[dvipsnames]{xcolor}
\usepackage{graphicx}
\usepackage{geometry}
\usepackage{booktabs}
\usepackage[T1]{fontenc}
\usepackage{fontawesome}
\usepackage{fancyhdr}
\usepackage{xspace}
\usepackage{enumitem}
\usepackage{etoc}
\usepackage{titletoc}
\usepackage{tcolorbox}
\usepackage{listings}

\usepackage{color}
\usepackage[dvipsnames]{xcolor}
\definecolor{JalapenoRed}{RGB}{183,21,64}
\definecolor{Belize}{RGB}{41,128,185}
\definecolor{Amour}{RGB}{238,82,83}
\usepackage{colortbl}

\usepackage{hyperref}

\newcommand{\takeaway}[2]{%
    \begin{tcolorbox}[
        colback=gray!10, colframe=teal!60!black, arc=5pt, boxrule=0.8pt,
        left=10pt, right=10pt, top=2pt, bottom=2pt, boxsep=5pt,
        enhanced jigsaw, drop shadow=gray!50!white,
        before skip=4pt, after skip=2pt
    ]
        \textbf{\textit{Takeaway #1:}} #2
    \end{tcolorbox}%
}

\usepackage[most]{tcolorbox}
\usepackage{enumitem}

\tcbset{
  promptbox/.style={
    title={#1},
    width=\columnwidth,
    breakable,
    enhanced,
    colback=gray!3,
    colframe=black!30,
    coltitle=black,
    fonttitle=\bfseries\footnotesize,
    boxrule=0.8pt,
    arc=1mm,
    boxsep=2pt,
    left=4pt,
    right=4pt,
    top=4pt,
    bottom=4pt,
    before skip=4pt,
    after skip=6pt
  }
}

\newcolumntype{L}[1]{>{\raggedright\arraybackslash}p{#1}}

\usepackage{titlesec}
\usepackage{titletoc}
\usepackage[most]{tcolorbox}
\usepackage{enumitem}

\usepackage[table]{xcolor}

\definecolor{GroundBG}{HTML}{6e8eb7}
\definecolor{ProfileBG}{HTML}{d79a77}
\definecolor{InteractBG}{HTML}{d77434}
\definecolor{OursBG}{HTML}{3e9b4e}

\newcommand{\Yes}{\makebox[1em][c]{\textcolor{green!55!black}{\ding{51}}}}
\newcommand{\Part}{\makebox[1em][c]{%
  \textcolor{orange!85!black}{%
    ~\ding{51}\hspace{-0.65em}\ding{55}
  }
}}
\newcommand{\No}{\makebox[1em][c]{\textcolor{red!75!black}{\ding{55}}}}

\usepackage[T1]{fontenc}
\usepackage[utf8]{inputenc}
\usepackage{microtype}
\usepackage{inconsolata}
\usepackage{graphicx}

\def\benchmark{\textbf{\texttt{{Trip+}}}\xspace}

\newcommand{\eg}{\textit{e}.\textit{g}.\xspace}

\usepackage[misc]{ifsym}
\usepackage{wrapfig}

\hypersetup{linkcolor=cyan}
\usepackage{mathtools}
\usepackage{ulem}
\usepackage{pifont}
\usepackage{bbding}

\usepackage{tcolorbox}
\usepackage{makecell}

\makeatletter
\let\orig@fnsymbol\@fnsymbol
\def\@fnsymbol#1{\ifcase#1\or\relax\else\orig@fnsymbol{#1}\fi}
\makeatother

\title{Trip+: Benchmarking Agents in Personalized Interactive Travel Planning}

\author{
\parbox{\textwidth}{
Junle Chen\textsuperscript{1}, \hspace{0.5mm}
Wei Chen\textsuperscript{1,2$~^{\textrm{\Letter}}$}, \hspace{0.5mm}
\textbf{Yehong Xu}\textsuperscript{1}, \hspace{0.5mm}
\textbf{Zhengjun Huang}\textsuperscript{1}, \hspace{0.5mm}
\textbf{Yuqian Wu}\textsuperscript{3}, \hspace{0.5mm}
\textbf{Zhoujin Tian}\textsuperscript{1}, \hspace{0.5mm}\\
\textbf{Kai Wang}\textsuperscript{2}, \hspace{0.5mm}
\textbf{Lei Wang}\textsuperscript{2}, \hspace{0.5mm}
\textbf{Xiaofang Zhou}\textsuperscript{1}
}
}

\affiliation{\textsuperscript{1}HKUST, \hspace{0.5mm} \textsuperscript{2}Tencent Hy, \hspace{0.5mm} \textsuperscript{3}HKUST(GZ)}

\abstract{
Interactive travel planning has become a popular use case for language models. Agents are deployed to manage evolving preferences and unexpected disruptions over multiple turns. Such settings require models to make complex, profile-conditioned planning decisions. However, existing benchmarks often evaluate feasibility, personalization, or interaction in relatively isolated settings. We therefore introduce \benchmark~to measure the ability of agents to plan travel holistically. In \benchmark, given traveler profiles and dynamic interactions, agents must generate and revise minute-level itineraries. End-to-end traveler experiences are evaluated via an LLM-based simulator, enabling the assessment of subjective metrics like fatigue. Our scenarios range from simple request resolutions to complex environment-driven replanning. We evaluate 18 LMs and find a consistent gap in experiential quality. Models favor technically feasible but exhausting itineraries that diverge sharply from profiled traveler preferences. 
}

\correspondence{jchenkg@connect.ust.hk, onedeanxxx@gmail.com, zxf@cse.ust.hk}
\repository{\url{https://github.com/junle-chen/trip-plus}}
\website{\url{https://junle-chen.github.io/trip-plus-site/}}
\date{\today}

\begin{document}

\maketitle

\makeatletter
\let\@fnsymbol\orig@fnsymbol
\makeatother

\section{Introduction}

As language agents~\citep{anthropic2026,openai2026,google2026gemini31pro,qwen2026qwen35} move toward real-world applications, travel planning~\citep{xie2024travelplanner} has emerged as a representative task that goes beyond one-shot execution. Unlike simple question answering, itinerary design naturally unfolds through multi-turn interactions: travelers refine preferences, introduce constraints, resolve conflicts, and react to changing travel conditions. This makes travel planning an ideal testbed for personalized agents, requiring them to maintain itineraries that are executable, profile-aligned, and consistent with accumulated user intents.


\begin{wrapfigure}{r}{7.5cm}
\begin{center}
\vspace{-6mm}
\includegraphics[width=1.0\linewidth]{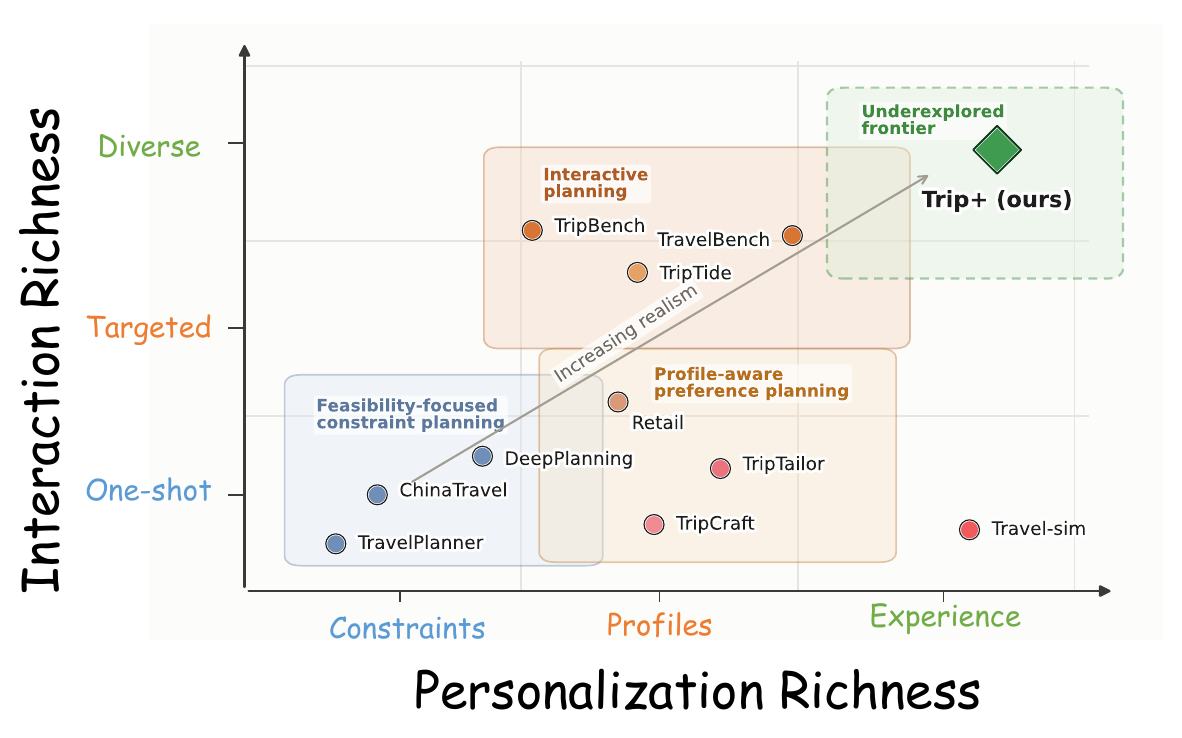}
\setlength{\abovecaptionskip}{-0.25cm}
\caption{Positioning travel-planning benchmarks for agents by personalization and interaction richness.} 
\label{fig.running_example}
\end{center}
\vspace{-2mm}
\end{wrapfigure} 
Figure~\ref{fig.running_example} illustrates the progressive evolution of existing travel-planning benchmarks along two dimensions: \textbf{\textit{personalization richness}} and \textbf{\textit{interaction richness}}. Along the personalization axis, early benchmarks emphasize explicit hard requirements and generic itinerary feasibility~\citep{xie2024travelplanner,shao2024chinatravel,zhang2026deepplanning}, while more recent work incorporates profile-conditioned preferences or traveler-experience evaluation~\citep{chaudhuri2025tripcraft,cheng2025travelbench,travelsim}. Along the interaction axis, benchmarks have progressed from one-shot itinerary generation to more interactive settings involving clarification, feedback incorporation, or replanning~\citep{cheng2025travelbench,shen2026trip,karmakar2025triptide}. These advances have substantially improved the realism of evaluating agents in travel-planning scenarios.

Despite this progress, as agents transition into real-world consumer products, an underexplored frontier remains: the joint handling of rich personalization and diverse long-horizon interactions. As shown in Table~\ref{tab:user_centered_travel_benchmarks}, existing travel-planning benchmarks fall short of comprehensive evaluation. \textit{Profile-aware benchmarks} typically evaluate static preference matching, often neglecting long-horizon interaction and stateful evaluation to test how agents handle evolving user intents. Conversely, \textit{interactive replanning benchmarks} frequently focus on isolated interaction patterns (\eg, asking clarifying questions, incorporating feedback, or performing isolated replanning), failing to comprehensively capture accumulated preference shifts, emerging conflicts, environmental disruptions, and their downstream impact on the profiled traveler's experience. This gap naturally raises a critical question: \textit{How can we evaluate whether travel-planning agents maintain feasible, traveler-suitable, and intent-consistent itineraries as user needs and travel conditions evolve across turns?}
    
\begin{table*}[t!]
\centering
\scriptsize
\setlength{\tabcolsep}{3.7pt}
\renewcommand{\arraystretch}{1.01}

\caption{
Comparison of travel-planning benchmarks across task construction, evaluation, and resource dimensions. Rows are grouped by their primary emphasis. \Yes{}, \Part{}, and \No{} denote explicit, partial/indirect, and no support, respectively. \textbf{Grounding}: environments integrated with a verifiable sandbox. \textbf{Profile}: inclusion of distinct user personas. \textbf{Interaction}: support for multi-turn user engagement. \textbf{Fine-grained}: requires detailed itineraries comprising diverse activities and transit, bounded by precise start/end timestamps. \textbf{Feasibility}: validates executability through structural completeness, entity grounding, strict temporal coherence, and accurate budget calculation. \textbf{Stateful Eval.}: tests the agent's ability to incorporate newly introduced constraints while strictly retaining prior user requirements across turns. \textbf{User Sim.}: leverages an LLM to role-play the profiled traveler, sequentially experiencing and scoring the fine-grained itinerary activities. \textbf{Open Source}: indicates whether data and evaluation code are publicly available. 
}

\label{tab:user_centered_travel_benchmarks}

\resizebox{\textwidth}{!}{
\begin{tabular}{@{}lcccc|ccc|c@{}}
\toprule
& \multicolumn{4}{c|}{\textbf{Task Construction}}
& \multicolumn{3}{c|}{\textbf{Evaluation}}
& \multicolumn{1}{c}{\textbf{Resources}} \\
\cmidrule(lr){2-5}
\cmidrule(lr){6-8}
\cmidrule(lr){9-9}
\textbf{Benchmark}
& \textbf{Grounding}
& \textbf{Profile}
& \textbf{Interaction}
& \textbf{Fine-grained}
& \textbf{Feasibility}
& \textbf{Stateful Eval.}
& \textbf{User Sim.}
& \textbf{Open Source} \\
\midrule

\rowcolor{gray!5}
\multicolumn{9}{@{}l@{}}{\textit{\textcolor{GroundBG}{Grounded feasibility}}} \\

\textsc{TravelPlanner}~\citep{xie2024travelplanner}
& \Yes & \No & \No & \No
& \Yes & \No & \No
& \Yes \\

\textsc{ChinaTravel}~\citep{shao2024chinatravel}
& \Yes & \No & \No & \Part
& \Yes & \No & \No
& \Yes \\

\textsc{DeepPlanning}~\citep{zhang2026deepplanning}
& \Yes & \No & \No & \Yes
& \Yes & \No & \No
& \Yes \\

\addlinespace[0.15em]
\rowcolor{gray!5}
\multicolumn{9}{@{}l@{}}{\textit{\textcolor{ProfileBG}{Profile-aware itinerary quality}}} \\

\textsc{TripCraft}~\citep{chaudhuri2025tripcraft}
& \Yes & \Yes & \No & \Yes
& \Yes & \No & \No
& \Part \\

\textsc{TripTailor}~\citep{wang2025triptailor}
& \Yes & \Part & \No & \Part
& \Yes & \No & \No
& \Yes \\

\textsc{Travel-Sim}~\citep{travelsim}
& \Yes & \Yes & \No & \Part
& \Part & \No & \Yes
& \No \\

\addlinespace[0.15em]
\rowcolor{gray!5}
\multicolumn{9}{@{}l@{}}{\textit{\textcolor{InteractBG}{Interactive replanning}}} \\

\textsc{RETAIL}~\citep{deng2025retail}
& \Yes & \Part & \Yes & \Part
& \Yes & \Part & \No
& \No \\

\textsc{TravelBench}~\citep{cheng2025travelbench}
& \Yes & \Yes & \Yes & \No
& \Yes & \Part & \No
& \Yes \\

\textsc{TripTide}~\citep{karmakar2025triptide}
& \Yes & \Part & \Yes & \Part
& \Yes & \Part & \No
& \No \\

\textsc{TRIP-Bench}~\citep{shen2026trip}
& \Yes & \Part & \Yes & \Part
& \Yes & \Part & \No
& \No \\

\midrule
\rowcolor{gray!10}
\textbf{\textcolor{OursBG}{\benchmark~(ours)}}
& \Yes & \Yes & \Yes & \Yes
& \Yes & \Yes & \Yes
& \Yes \\

\bottomrule
\end{tabular}
}
\vspace{-4mm}
\end{table*}

To answer this question, we introduce \benchmark, a real-world travel-planning benchmark specifically designed to evaluate language agents in personalized, multi-turn environments. \benchmark~grounds 11 distinct traveler profiles within a 40-city evidence sandbox, accessible via 11 domain-specific tools (covering attractions, restaurants, hotels, mobility, intercity transport, weather, and location evidence). Comprising 153 multi-turn instances and 570 user turns, the benchmark captures four realistic interaction archetypes: \textit{User-State Evolution}, \textit{Request Resolution}, \textit{Environment-Driven Replanning}, and \textit{Long-Horizon Alignment}. At each turn, an agent must strategically navigate its action space—outputting a \textsc{Plan}, requesting \textsc{Clarification} for conflicting or missing information, or returning \textsc{NoSolution} when tool evidence proves hard constraints are unsatisfiable. To rigorously assess these actions, \benchmark~employs a comprehensive four-layer evaluation protocol. It first validates the correctness of the chosen response mode (gating), followed by deterministic checks for itinerary executability. Stateful rule-based evaluators then verify adherence to requirement satisfaction and cross-turn intent retention, while a profile-conditioned large language model simulator finally replays the minute-level execution trace to score the holistic traveler experience (\eg, fatigue, pacing pressure, weather exposure, budget stress, and preference mismatch). A sampled subset has been reviewed by human experts to ensure the reliability of judgments.
In summary, our contributions are threefold:

\begin{itemize}[leftmargin=*,itemsep=0.2em]
    \item \textit{Task Construction.} We introduce \benchmark, a personalized multi-turn travel-planning benchmark for frontier language agent model, comprising 153 instances and 570 user turns. It is built from 11 traveler-profile templates and a fixed 40-city evidence sandbox exposed through 11 domain-specific travel tools, covering four realistic interaction archetypes.
    \item \textit{Evaluation Setup.} We propose a profile- and interaction-aware evaluation protocol that assesses response mode, feasibility, hard constraints, profile alignment, cross-turn retention, and simulated traveler experience.
    \item \textit{Findings.} Our studies indicate that mere feasibility is insufficient for travel-planning: current agents often underperform in intention retention, profile matching, and traveler experience across diverse interactions.
\end{itemize}

\section{\benchmark: Towards Benchmarking Product Agents for Real Users}
\label{sec:usertrip}

\begin{figure*}[htbp!]
    \centering
    \includegraphics[width=\textwidth]{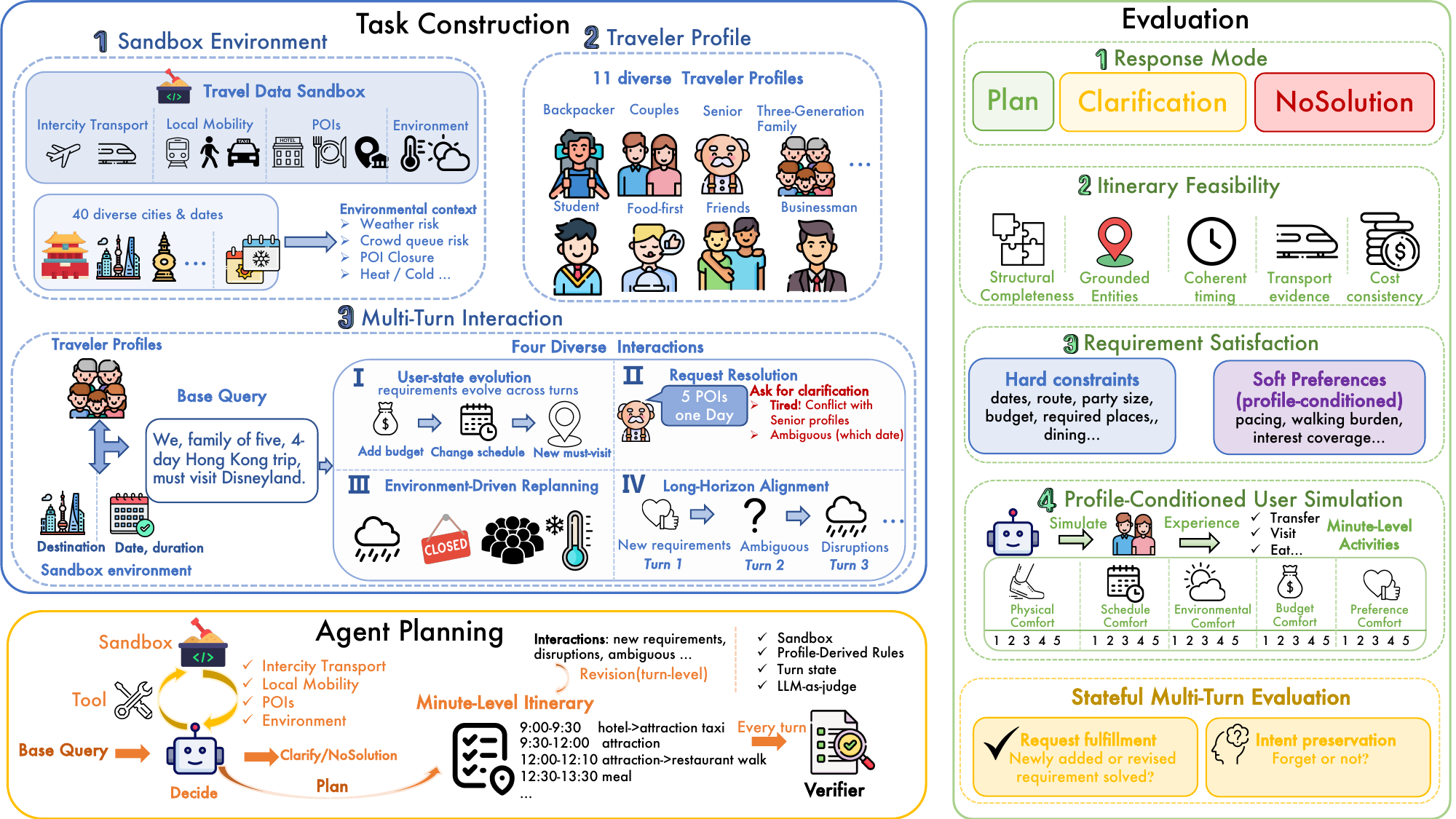}
    \caption{
    The overall design of \textsc{\benchmark}, used to evaluate language agents in complex travel planning scenarios.
    }
    \label{fig:overview}
\end{figure*}

This section describes \benchmark in detail (Figure~\ref{fig:overview}).
We first formalize the task (\secref{sec:task-formulation}),
then explain how task instance is constructed through a sandbox, traveler profiles, and multi-turn interaction (\secref{sec:benchmark-construction}). Besides, we review benchmark statistics (\secref{sec:dataset}). Finally, we introduce our evaluation design (\secref{sec:evaluation-protocol}).

\subsection{Task Formulation}
\label{sec:task-formulation}
\benchmark evaluates multi-turn travel-planning agents in a fixed travel sandbox.
At each turn, the agent uses the visible traveler profile, the dialogue prefix, and tool-retrieved sandbox evidence to choose one of three response modes: generate a minute-level itinerary, ask for clarification, or return an evidence-backed no-solution response.
The benchmark also maintains hidden turn states only for evaluation.

\textbf{Instance.}
We formalize each task instance as an agent-visible part and a hidden reference part:
\[
\begin{gathered}
x=(x^{\mathrm{vis}},x^{\mathrm{ref}}),\\
x^{\mathrm{vis}}=(\mathcal{D}^{\mathrm{tool}},p^{\mathrm{vis}},I_{1:T}),\\
x^{\mathrm{ref}}=(\mathcal{D}^{\mathrm{ref}},s_{1:T}).
\end{gathered}
\]
The visible part \(x^{\mathrm{vis}}\) consists of the tool-accessible sandbox \(\mathcal{D}^{\mathrm{tool}}\), the visible traveler profile \(p^{\mathrm{vis}}\), and turn-level interaction histories \(I_{1:T}=(I_1,\ldots,I_T)\). 
Each \(I_t=(u_1,\hat{r}_1,\ldots,u_{t-1},\hat{r}_{t-1},u_t)\) is the dialogue prefix up to turn \(t\), where \(u_i\) and \(\hat{r}_i\) denote the user utterance and agent response at turn \(i\).

The hidden reference part \(x^{\mathrm{ref}}\) contains two evaluation resources: 
\(\mathcal{D}^{\mathrm{ref}}\), which grounds itinerary-feasibility checks in the fixed sandbox evidence, and 
\(s_{1:T}=(s_1,\ldots,s_T)\), a sequence of hidden states aligned with the \(T\)-turn dialogue.

\textbf{Hidden state.}
Each turn \(t\) has a hidden state:
\[
s_t=(m_t,\mathcal{H}_t,\mathcal{Q}_t,\mathcal{E}_t),
\]
where \(m_t\) is the expected response mode, 
\(\mathcal{H}_t\) contains hard constraints that apply at turn \(t\), 
\(\mathcal{Q}_t\) contains profile-derived soft preferences relevant to turn \(t\), 
and \(\mathcal{E}_t\) contains environmental conditions that affect planning.
These fields define the turn-level evaluation targets: response-mode correctness, hard-constraint satisfaction, soft-preference satisfaction, and environment-aware planning.

Across turns, \benchmark~updates the hard constraints, profile-derived soft preferences, and environmental conditions that remain relevant to the dialogue:
\[
\begin{aligned}
\mathcal{H}_t &=
(\mathcal{H}_{t-1}\cup\Delta_t^{H,+})\setminus\Delta_t^{H,-},\\
\mathcal{Q}_t &=
(\mathcal{Q}_{t-1}\cup\Delta_t^{Q,+})\setminus\Delta_t^{Q,-},\\
\mathcal{E}_t &=
(\mathcal{E}_{t-1}\cup\Delta_t^{E,+})\setminus\Delta_t^{E,-}.
\end{aligned}
\]
Here, \(\Delta_t^{\cdot,+}\) denotes newly introduced items, while \(\Delta_t^{\cdot,-}\) denotes items that are revised or canceled.
Items not removed remain active in later turns, enabling evaluation of cross-turn retention.

\textbf{Turn-level agent behavior.}
At turn \(t\), the agent \(\mathcal{A}_\theta\) observes the visible context, including the traveler profile \(p^{\mathrm{vis}}\), the interaction histories \(I_{\leq t}\), and sandbox evidence \(\mathcal{R}_t\) retrieved through tool calls. 
It returns a response with two components: a response mode $\hat{m}_t$ and mode-specific content $\hat{z}_t$:
\[
\begin{aligned}
\hat{r}_t
&=(\hat{m}_t,\hat{z}_t)
= \mathcal{A}_\theta(p^{\mathrm{vis}}, I_{\leq t}, \mathcal{R}_t),
\quad
\mathcal{R}_t \subseteq \mathcal{D}^{\mathrm{tool}},
 \\
\hat{z}_t
&=
\begin{cases}
\hat{y}_t, & \hat{m}_t=\mathrm{Plan},\\
\eta_t, & \hat{m}_t\in\{\mathrm{Clar},\mathrm{NoSol}\}.
\end{cases}
\end{aligned}
\]
Here, \(\eta_t\) denotes a non-plan response: a clarification question under \(\mathrm{Clar}\), or an infeasibility explanation under \(\mathrm{NoSol}\). 
For plan responses, \(\hat{y}_t\) is an executable itinerary organized by day:
\[
\begin{aligned}
\hat{y}_t &= (d_1,\ldots,d_D,B), 
\quad d_j=(c_j,A_j,h_j), \\
A_j &= (a_{j1},\ldots,a_{jn_j}),
\quad j=1,\ldots,D.
\end{aligned}
\]
Here, \(B\) is the itinerary-level budget summary; \(c_j\), \(A_j\), and \(h_j\) denote the city, ordered activities, and accommodation for day \(j\). 
Each activity \(a_{jk}\) records its time interval, type, grounded place or transport item, movement context, and cost fields. 
The benchmark evaluates both the selected mode \(\hat{m}_t\) and the content \(\hat{z}_t\) against the hidden state \(s_t\).

\subsection{Benchmark Construction}
\label{sec:benchmark-construction}

\benchmark~is constructed in three stages: building a fixed travel data sandbox, defining traveler profiles, and generating multi-turn interactions, making travel planning realistic and verifiable. 

\textbf{Sandbox construction.}
We build a fixed travel sandbox over a diverse set of 40 Chinese cities, covering different destination types, seasons, and local travel conditions. 
The sandbox provides reproducible evidence for POIs, hotels, restaurants, weather, local mobility, and intercity transport, together with city-level environmental context for environment-aware scenarios. 
Agents access this evidence through 11 domain-specific travel tools, while evaluators use the same fixed snapshot to verify grounding, feasibility, costs, and environment-aware planning. 
Detailed evidence fields and coverage are reported in Appendix~\ref{app:sandbox}, ~\ref{app:tool-interface} and ~\ref{app:environment-references}.

\textbf{Traveler profile construction.}
We construct 11 diverse traveler-profile templates to systematically vary long-term user context across planning instances. The templates are designed to cover heterogeneous traveler needs, including party composition, budget sensitivity, mobility constraints, pace, interests, accommodation preferences, and travel style. 
For each instance, we sample an observable profile \(p^{\mathrm{vis}}\) and provide it to the agent as user memory. 
Each observable profile is also mapped to hidden profile-derived rules used only for evaluation, allowing \benchmark~to test whether agents produce plans that are not only feasible but also suitable for the target traveler. 
Detailed profile fields and derived rules are provided in Appendix~\ref{app:profile-schema}.

\begin{figure*}[!t]
    \centering
    \includegraphics[width=\textwidth]{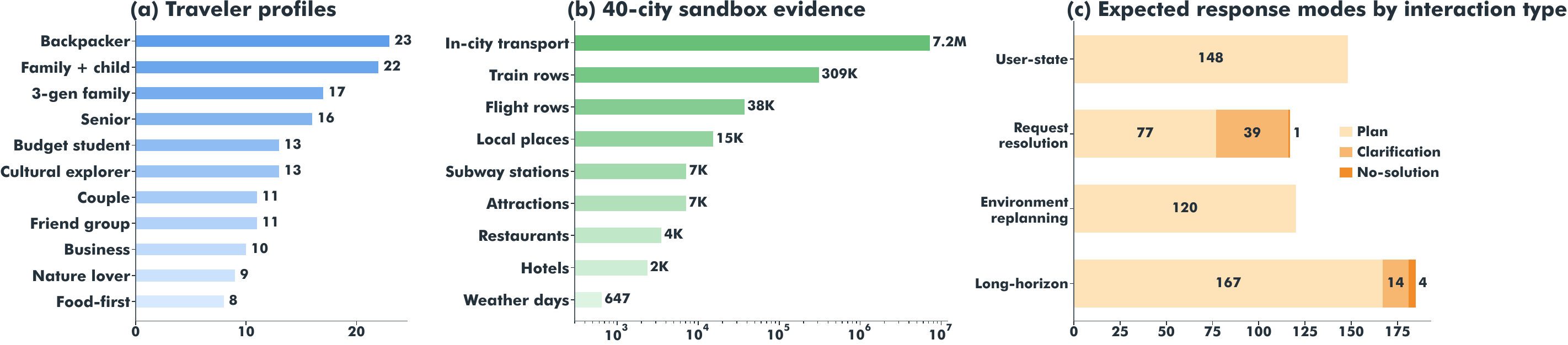}
    \caption{
    Dataset statistics of \benchmark, covering traveler profiles, sandbox evidence, and multi-turn interactions.
    }
    \label{fig:dataset_stat}
\end{figure*}


\textbf{Multi-turn interaction construction.}
We generate multi-turn interactions using a state-first pipeline. For each turn, we specify a structured hidden state (including state delta, response mode, and evaluation target) before rendering it as a natural-language user utterance. Specifically:

\begin{itemize}[leftmargin=*,itemsep=0.2em]
\item \textit{Sample a sandbox-grounded seed request.}
We first sample an executable trip frame from the fixed evidence sandbox, including the route, dates, trip duration, party size, room count, and intercity transport mode. We then sample two visible database-backed requirement categories from transport, lodging, dining, and attractions, with budget added as an extra visible constraint when it fits the sampled traveler.

\item \textit{Expand into a multi-turn scenario.} We expand each seed request using one of four interaction archetypes: \textit{\uppercase\expandafter{\romannumeral1}. User-state Evolution}, where user needs (\eg, party composition, budget, schedule) change across turns; \textit{\uppercase\expandafter{\romannumeral2}. Request Resolution}, where the agent clarifies or resolves underspecified, conflicting, or infeasible requests; \textit{\uppercase\expandafter{\romannumeral3}. Environment-driven Replanning}, where external disruptions (\eg, weather, closures, traffic) require itinerary revisions; \textit{\uppercase\expandafter{\romannumeral4}. Long-horizon Alignment}, where multiple updates are handled while preserving prior commitments. For each turn, we maintain a hidden state $s_t$ tracking the expected response mode, active hard constraints, accumulated user updates, and active environment events. We also store the turn delta and preservation targets to specify current turn changes and the prior commitments that must remain valid.

\item \textit{Render and validate each turn.}
Once the hidden turn state and update are fixed, we convert them into the actual user message shown to the agent.
Here, we keep the required entities, numbers, constraints, and response-mode cues unchanged, discarding any rendered message that omits key details or exposes hidden reference fields.

\end{itemize}


\subsection{Dataset Statistics} \label{sec:dataset}
\benchmark~contains 153 multi-turn instances and 570 user turns. Figure~\ref{fig:dataset_stat} summarizes the profile, sandbox evidence, and interaction coverage. The instances span all 11 traveler profiles (8--23 instances per profile), grounded in a fixed 40-city sandbox containing 7.2M in-city transport records, 309K train rows, 38K flight rows, and thousands of places, attractions, stations, restaurants, hotels, and weather records. The turns cover four interaction archetypes and all three response modes; planning dominates, while clarification and no-solution cases appear mainly in request-resolution and long-horizon interactions. More distributions of trip duration and departure month are in Appendix~\ref{app:sandbox}.

\subsection{Evaluation Protocol} \label{sec:evaluation-protocol}

Each turn in \benchmark is evaluated against its hidden turn state $s_t$, which specifies the expected response mode $m_t$, active hard requirements $\mathcal{H}_t$, profile-derived expectations $\mathcal{Q}_t$, environment conditions $\mathcal{E}_t$, and items to preserve across turns.
Evaluation proceeds in two stages: first, we check if the response follows the expected mode; for planning turns that pass this mode gate, the remaining evaluation covers operational feasibility, turn-state satisfaction, profile-conditioned user simulation, and stateful multi-turn evaluation. Specifically:

\textbf{Response-Mode Gating.}
The mode gate determines whether an agent's response is eligible for itinerary-level evaluation. If the response mode mismatches $m_t$, we record a mode error and skip itinerary-level metrics. Itinerary metrics are computed only when $m_t=\textsc{Plan}$ and the response provides a concrete itinerary; \textsc{Clarification} and \textsc{NoSolution} turns are evaluated by response-mode accuracy instead of itinerary quality.

\textbf{Itinerary Feasibility.}
Itinerary feasibility measures whether a generated itinerary is executable under the fixed travel sandbox. The evaluator checks structural completeness, grounded choices (hotels, restaurants, attractions, transport), temporal coherence, venue opening hours, supported transfers, and cost arithmetic.

\textbf{Requirement Satisfaction.}
This part evaluates whether an executable itinerary satisfies the active user state at turn $t$. We separately measure \textit{hard-constraint satisfaction} and \textit{soft-preference satisfaction}. Hard constraints in $\mathcal{H}_t$ correspond to explicit user requirements (\eg, dates, destinations, party size, budget, required lodging/dining/transport); violations indicate request failures. Soft preferences in $\mathcal{Q}_t$ capture profile-derived expectations (\eg, pace, walking tolerance, budget sensitivity, comfort, interests). Both are evaluated by deterministic rules, treating hard violations as request failures and soft violations as profile-alignment failures.

\textbf{Profile-Conditioned User Simulation}.
Rule-based checks verify whether the generated itinerary $\hat{y}_t$ is feasible, satisfies hard constraints, and aligns with verifiable soft preferences. However, they cannot model how a profiled traveler subjectively experiences an itinerary activity by activity, as an otherwise valid plan may still feel tiring, rushed, or uncomfortable. To capture this, we feed the activity sequences $\{A_j\}_{j=1}^{D}$ in $\hat{y}_t$ to an LLM simulator that evaluates each activity from the traveler's perspective under active profile and environmental conditions. It assigns 1--5 scores with brief rationales across applicable dimensions (physical, schedule, environmental, budget comfort, and preference satisfaction), complementing rule-based metrics by assessing subjective experiential suitability.

\textbf{Stateful Multi-Turn Evaluation.}
Multi-turn evaluation is stateful, judging each response against hidden turn states rather than in isolation. The transition from $s_{t-1}$ to $s_t$ defines the current turn changes, while $s_t$ specifies which hard constraints, soft preferences, and environment conditions remain active. We report two signals: \textit{Request fulfillment} verifies whether the response incorporates new turn changes (added or revised hard constraints $\Delta_t^{H}$, soft preferences $\Delta_t^{Q}$, and environment conditions $\Delta_t^{E}$); \textit{Intent preservation} checks whether ongoing constraints, preferences, and environment conditions in $\mathcal{H}_t$, $\mathcal{Q}_t$, and $\mathcal{E}_t$ remain satisfied. Together, they measure the agent's ability to adapt to updates while retaining prior commitments.

\noindent\underline{Note, full metric definitions are in Appendix~\ref{app:evaluation-metrics}.}

\section{Experiments}


\subsection{Experimental Setup}

\paragraph{Models and Agent Scaffold.}
We evaluate 18 Agentic models across frontier families including Gemini, GPT, Doubao, DeepSeek, MiniMax, Kimi, Hy, and GLM, as well as lightweight families including Gemma, Qwen, and GLM. 
Exact model variants are reported in Appendix~\ref{app:model} and ~\ref{app:model_p}.
To focus on the models' own tool-use and planning capabilities, all models use the same lightweight OpenAI-compatible function-calling scaffold.

\paragraph{Metrics.}
We report two groups of metrics. 
\textit{End-to-End Interaction} measures turn-level reliability over all interactions, including response-mode accuracy, request fulfillment, and intent preservation. 
\textit{Valid Plan Quality} evaluates plan outputs that pass the response-mode gate, covering operational feasibility, hard-constraint satisfaction, profile-conditioned soft preferences and simulated traveler experience. 
For simulated traveler experience, we use four fixed profile-conditioned judges from the Qwen, Claude, Gemini, and GPT families to score each generated plan from the target traveler's perspective, and take the median as the final value. 
Appendix~\ref{app:user_sim_reliability} lists the exact judge versions and reports reliability analyses: pairwise Spearman rank alignment, ensemble-level Cronbach's \(\alpha\) and human rationale verification over 50 sampled cases covering 1,825 activities.

\begin{table*}[t]
\centering
\scriptsize
\setlength{\tabcolsep}{3.0pt}
\renewcommand{\arraystretch}{1.2}

\definecolor{FrontierHead}{HTML}{D8E7FF}
\definecolor{LightHead}{HTML}{DDEFD8}

\providecommand{\best}[1]{\textbf{#1}}
\providecommand{\second}[1]{\underline{#1}}

\caption{
Main results on \benchmark. 
Models are grouped into frontier and lightweight models. 
Bold and underline denote the best and second-best scores in each column.
\textbf{Plan Avg.} averages the four valid-plan quality metrics.
\textbf{Win(\%)} reports the percentage of non-aggregate metrics on which each model ranks first.
}
\label{tab:main_results}

\resizebox{\textwidth}{!}{
\begin{tabular}{@{}lccc|cccc|cc@{}}
\toprule
& \multicolumn{3}{c|}{\textbf{End-to-End Interaction}}
& \multicolumn{4}{c|}{\textbf{Valid Plan Quality}}
& \multicolumn{2}{c}{\textbf{Summary}} \\
\cmidrule(lr){2-4}
\cmidrule(lr){5-8}
\cmidrule(l){9-10}
\textbf{Model}
& \textbf{Resp. Acc.}
& \textbf{Req. Fulfill}
& \textbf{Intent Pres.}
& \textbf{Feas.}
& \textbf{Hard Cons.}
& \textbf{Soft Pref.}
& \textbf{User Sim.}
& \textbf{Plan Avg.}
& \textbf{Win(\%)} \\
\midrule

\rowcolor{FrontierHead}
\multicolumn{10}{@{}l}{\textit{Frontier models}} \\

Gemini-3.1-Pro-Preview      
& 0.8544 & \second{0.7382} & \best{0.7697}
& \best{0.8812} & \best{0.9047} & \second{0.6389} & 0.5077
& \best{0.7331} & \best{42.9} \\

Gemini-3-Flash-Preview      
& 0.8667 & 0.6885 & 0.6532
& 0.7099 & 0.7179 & 0.6172 & 0.4862
& 0.6328 & 0.0 \\

GPT-5.4                     
& 0.7544 & 0.5585 & 0.5489
& 0.7669 & 0.8110 & 0.5854 & \best{0.5184}
& 0.6704 & 14.3 \\

GPT-5.4-Mini                
& 0.7807 & 0.4879 & 0.4221
& 0.6455 & 0.6126 & 0.6207 & 0.5085
& 0.5968 & 0.0 \\

GLM-5.1                     
& 0.8702 & 0.7214 & 0.7301
& \second{0.8463} & 0.7973 & 0.6136 & 0.5014
& \second{0.6896} & 0.0 \\

Kimi-K2.6                   
& 0.8439 & 0.6727 & 0.6811
& 0.7472 & 0.7984 & 0.6380 & 0.5164
& 0.6750 & 0.0 \\

Doubao-Seed-2.0-Pro          
& 0.8246 & 0.6811 & 0.6814
& 0.6731 & 0.8006 & 0.6031 & 0.5128
& 0.6474 & 0.0 \\

DeepSeek-V4-Pro             
& 0.7912 & 0.6657 & 0.7139
& 0.6755 & \second{0.8375} & 0.6258 & 0.5100
& 0.6622 & 0.0 \\

DeepSeek-V3.2               
& 0.7702 & 0.5418 & 0.5566
& 0.6461 & 0.7162 & 0.5842 & 0.4755
& 0.6055 & 0.0 \\

Hy3-preview
& \second{0.8930} & 0.6708 & \second{0.7640}
& 0.7223 & 0.6476 & 0.5957 & 0.4866 
& 0.6130 & 0.0 \\

MiniMax-M2.7                
& 0.7772 & 0.5766 & 0.4921
& 0.5596 & 0.6326 & 0.6076 & 0.4992
& 0.5748 & 0.0 \\

\midrule

\rowcolor{LightHead}
\multicolumn{10}{@{}l}{\textit{Lightweight models}} \\

Gemma-4-31B                 
& \best{0.9035} & \best{0.7465} & 0.7490
& 0.7427 & 0.7632 & 0.5731 & 0.4923
& 0.6428 & \second{28.6} \\

Gemma-4-26B-A4B             
& 0.8386 & 0.5989 & 0.5594
& 0.6761 & 0.6925 & 0.5940 & 0.4916
& 0.6136 & 0.0 \\

Qwen3.5-27B                 
& 0.8491 & 0.6309 & 0.6218
& 0.7393 & 0.7256 & 0.6375 & 0.4821
& 0.6461 & 0.0 \\


%

Qwen3.5-122B-A10B-FP8        
& 0.3368 & 0.2033 & 0.1028
& 0.6935 & 0.7235 & 0.6319 & 0.5050
& 0.6385 & 0.0 \\

Qwen3.6-27B                 
& 0.5649 & 0.4350 & 0.3756
& 0.7521 & 0.7225 & \best{0.6626} & \second{0.5167}
& 0.6635 & 14.3 \\

Qwen3.6-35B-A3B             
& 0.6912 & 0.4554 & 0.2463
& 0.4292 & 0.4588 & 0.5839 & 0.5141
& 0.4965 & 0.0 \\

GLM-4-32B                   
& 0.5053 & 0.4656 & 0.3155
& 0.5236 & 0.5605 & 0.6141 & 0.4791
& 0.5443 & 0.0 \\

\bottomrule
\end{tabular}
}
\end{table*}

\begin{figure*}[t]
\centering
\includegraphics[width=\textwidth]{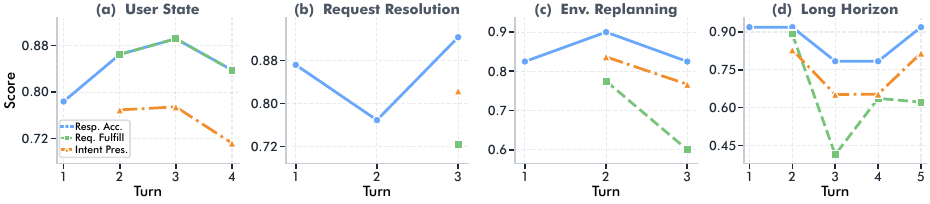}
\caption{
Interaction reliability of Gemini-3.1-Pro-Preview across four multi-turn scenario types.
Intent preservation is omitted at Turn 1 because there is no previous commitment to preserve.
}
\vspace{-3mm}
\label{fig:interaction_scenarios}
\end{figure*}

\subsection{Main Results} \label{sec:main_result}

\noindent\ding{182}\
\textit{Current LLM agents remain unreliable in realistic multi-turn travel planning.}
Table~\ref{tab:main_results} shows clear weaknesses in the interaction: models still make mistakes in deciding whether to plan, ask for clarification, or report infeasibility, and they often fail to preserve earlier user needs as the dialogue evolves. 
Gemma-4-31B achieves the best response accuracy and request fulfillment, while Gemini-3.1-Pro-Preview performs best on intent preservation. 
These results suggest that diverse user and environment changes expose interaction-level failures that are not captured by static planning alone.
\noindent\ding{183}\
\textit{Feasible itineraries are not necessarily user-aligned.}
Gemini-3.1-Pro-Preview achieves the strongest valid-plan quality, ranking highest in feasibility, hard-constraint satisfaction and plan average score. 
However, softer alignment metrics remain weak: soft-preference scores are lower, and the best user-simulation score achieved by GPT-5.4 is only 0.5184. 
This shows that satisfying explicit constraints is still insufficient for matching users' implicit preferences and evolving expectations.


\section{In-depth Analysis}
\label{sec:interaction_reliability}
Building on the main results in Section~\ref{sec:main_result}, we focus on \textit{Gemini-3.1-Pro-Preview}, the strongest overall model in Table~\ref{tab:main_results}, to diagnose two remaining gaps: \textit{unreliable interaction and unsuitable plans}. 
We also analyze inference cost to test whether more computation improves planning performance.

\begin{figure*}[t]
    \centering
    \includegraphics[width=\textwidth]{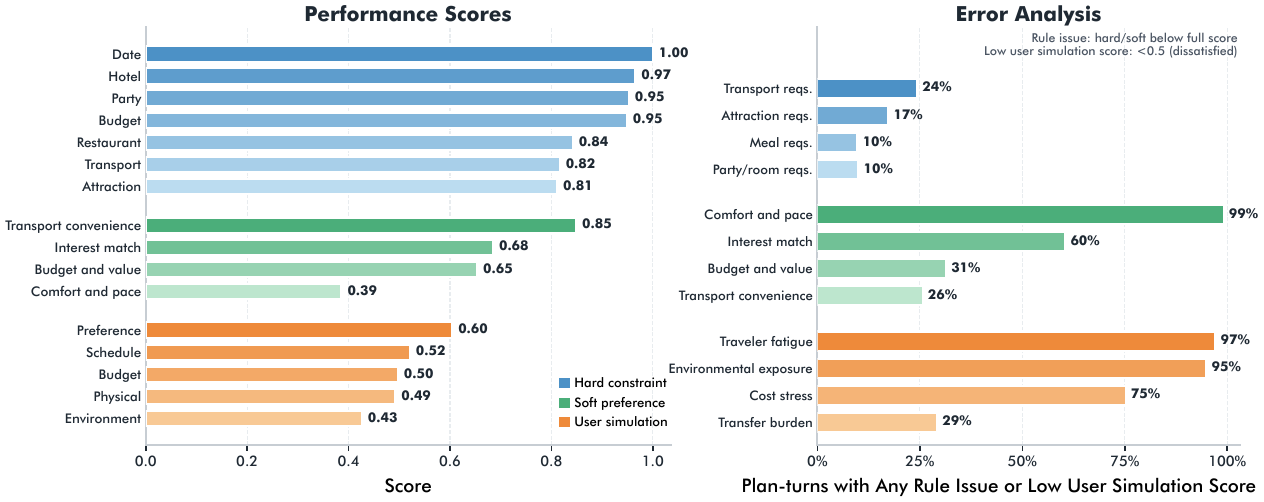}
    \caption{Performance scores and error analysis of Gemini-3.1-Pro-Preview generated plans}
    \label{fig:error_analysis}
\end{figure*}

\subsection{Interaction Reliability across Scenarios}
Figure~\ref{fig:interaction_scenarios} shows that unreliable interaction mainly comes from state-consistent revision rather than response-mode selection across all four scenarios.
\noindent\ding{182}\ \textit{The model handles early updates, but struggles as requirements accumulate.}
Response accuracy and request fulfillment initially improve from Turn 1 to 3, indicating that the model can initially incorporate user-state updates. However, as requirements accumulate, request fulfillment drops at Turn 4 and intent preservation falls to 0.71, showing that the model struggles to satisfy the growing set of requirements and begins to forget earlier user requests.
\ding{183} \textit{Clarification remains a bottleneck in request resolution.}
At the ambiguous Turn 2, Response Accuracy drops temporarily to 0.77, showing that the model is less reliable when clarification is required. After the ambiguity is resolved, response accuracy recovers, but final request fulfillment reaches only 0.72. This gap shows that the model can return to the correct response mode after clarification, but still struggles to apply the clarified requirement when revising the itinerary.
\noindent\ding{184}\ \textit{The model detects environment changes, but often revises plans incorrectly.}
In environment-driven replanning, response accuracy reaches 0.90, indicating that the model reliably detects when to update the itinerary. However, Request Fulfillment drops to 0.60 and Intent Preservation to 0.77, demonstrating a clear deficiency in producing revised plans that satisfy new conditions while preserving past constraints.
\noindent\ding{185}\ \textit{Long-horizon alignment is the hardest scenario.}
Under accumulated constraints, the ambiguous Turn 3 triggers a severe performance bottleneck: Response Accuracy drops to 0.78, while Request Fulfillment plummets sharply to 0.41, highlighting the model's high vulnerability to ambiguous inputs as constraints grow.

\vspace{2mm}
\takeaway{1}{Better interaction requires revising plans reliably as constraints grow.}

\subsection{Traveler Suitability and Failure Modes}
\label{sec:personalization}

We further analyze valid plans generated by Gemini-3.1-Pro-Preview. 
Figure~\ref{fig:error_analysis} reports both component scores and error rates. We observe:
\ding{182} \textit{Hard constraints are strong overall, but itinerary-level choices remain the main failure source.} 
Dates, hotels, party size, and budget are handled reliably, all scoring at or above 0.95. 
However, constraints that require concrete itinerary decisions are weaker: restaurants, transport, and attractions score 0.84, 0.82, and 0.81, respectively. 
These lower scores are reflected in the error rates, with transport, attraction, and meal requirement errors appearing in 24\%, 17\%, and 10\% of valid plans.
\noindent\ding{183}\ \textit{Pace is the most systematic personalization failure.}
Among soft preferences, comfort and pace receives the lowest score at 0.39. 
The error analysis shows the same pattern: 99\% of valid plans contain pace-related burden, such as overly dense schedules, insufficient rest, or tight transitions. 
This indicates that many feasible itineraries are still exhausting or poorly paced for the target traveler.
\noindent\ding{184}\ \textit{Valid plans still impose substantial experience burden.}
Even among valid plans, user-simulation scores fall below 0.5 on key experience dimensions, including environment, schedule and physical burden. 
The error analysis shows that these problems are frequent: traveler fatigue appears in 97\% of valid plans, and environmental exposure appears in 95\%. 
This shows that many valid itineraries are still tiring or environmentally unsuitable for the target traveler.

\vspace{2mm}
\takeaway{2}{Better personalization depends on pacing and burden control.}

\subsection{Cost–Performance Trade-off}


\begin{wrapfigure}{r}{8cm}
\begin{center}
\vspace{-9mm}
\includegraphics[width=1.0\linewidth]{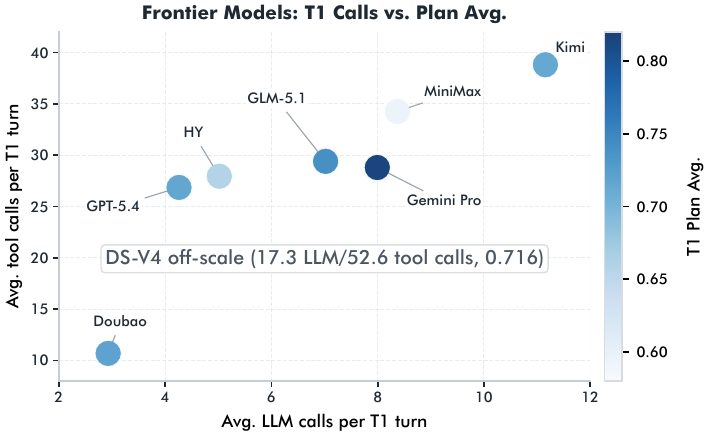}
\setlength{\abovecaptionskip}{-0.25cm}
\caption{Turn-1 Inference Cost vs. Plan Average.} 
\label{fig.cost_performance}
\end{center}
\end{wrapfigure} 
We analyze whether higher inference effort improves Turn-1 Plan Avg.
We use the average number of LLM calls and tool calls at Turn 1, because the first turn carries the full planning burden of building the itinerary, while later turns mainly revise it. 
Figure~\ref{fig.cost_performance} shows no clear monotonic relationship between cost (LLM/tool calls) and Plan Avg. among frontier models: DS-V4 and Kimi use more calls, while Gemini Pro achieves the highest Turn-1 Plan Avg. with moderate usage. 
We report the task-level cost analysis across benchmark metrics in Appendix ~\ref{app:inference}, which shows the similar pattern. 
Together, these results suggest that \textit{effective evidence use and plan construction matter more than simply increasing the number of calls.}

\vspace{2mm}
\takeaway{3}{Better planning depends on effective evidence use, not more calls.}

\section{Related Work}


\textbf{Travel Planning Benchmarks.} Travel planning is a prime testbed for language agents. Early benchmarks primarily focused on basic feasibility and structured planning~\citep{xie2024travelplanner,shao2024chinatravel,zhang2026deepplanning}. Later works expanded into fine-grained itinerary quality, personalization, and complex interactive scenarios like disruption-driven replanning~\citep{chaudhuri2025tripcraft,wang2025triptailor,travelsim,deng2025retail,cheng2025travelbench,shen2026trip,karmakar2025triptide}. Despite this progress, existing evaluations lack a unified framework for dynamic contexts and granular experiential feedback. In contrast, \benchmark~differentiates itself by treating the active user state, profile-derived suitability rules, expected response mode, and activity-level experience trace as jointly constructed oracle fields.


\textbf{User-Centered Interactive Evaluation.}
Recent studies~\cite{laban2025llms} indicate that as agents deploy into the real world, language models notoriously "get lost in multi-turn conversations," failing to track long-term user states. To address this, recent benchmarks shift to interactive evaluation: \textsc{UserBench} tests preference elicitation with simulated users, while $\tau$-bench evaluates tool-mediated interactions under policy constraints~\citep{qian2025userbench,yao2024tau}. While these works establish baseline capabilities in conversation tracking, they fall short in physical-world coordination. \benchmark~extends this perspective to travel planning, challenging agents to remain unlost across executable timed itineraries, dynamic environmental changes, and personalized traveler profiles.

\section{Conclusion}

We introduce \benchmark{}, a benchmark evaluating travel-planning agents on generating feasible, traveler-suitable, and intent-consistent itineraries under dynamic user needs and environments. Our evaluation reveals a clear gap: while current models satisfy basic feasibility and explicit constraints, they struggle significantly with stateful revisions, user alignment, and effective evidence use. Ultimately, we hope \benchmark{} drives the development of agents that plan adaptive, profile-aligned experiences rather than merely executable trips.

\clearpage


\section*{Limitations}

While \benchmark~provides a realistic setup, it has three primary limitations. First, synthetic queries lack organic dialogue noise. We utilize them to ensure a reproducible baseline, leaving authentic conversation integration for future work. Second, fixed evaluation rules may suffer from incomplete coverage, a challenge we leave for future research on dynamic rule generation. Finally, the LLM as a judge paradigm cannot capture genuine physical or emotional experiences. We mitigate this using multiple models and human verification, though authentic human testing remains a crucial next step.

\section*{Ethical Considerations}

\benchmark~is designed strictly for research. It relies on a fixed sandbox and synthetic profiles, posing no privacy risks. To prevent bias, we treat traveler profiles solely as objective planning constraints rather than demographic stereotypes. The sandbox mimics actual travel entities, but these are reproducible references rather than live availability or safety guarantees. Plus, our user simulations serve only as diagnostic signals for itinerary quality and should never substitute for actual human feedback.

\definecolor{textgray}{HTML}{6E6E73}
\makeatletter
\newcommand\applefootnote[1]{%
  \begingroup
  \renewcommand\thefootnote{}%
  \renewcommand\@makefntext[1]{\noindent##1}%
  \footnote{#1}%
  \addtocounter{footnote}{-1}%
  \endgroup
}
\makeatother

\bibliography{ref}
\bibliographystyle{plainnat}
\newpage
\appendix

\clearpage
\begin{center}
    \Large{\sc\Huge Appendix\\ \normalsize \benchmark: Benchmarking Agents in Personalized Interactive Travel Planning}\\
\end{center}
\vskip 4mm
\startcontents[sections]
\vbox{\sc\Large Table of Contents}
\vspace{5mm}
\hrule height .8pt
\vspace{-2mm}
\printcontents[sections]{}{1}{\setcounter{tocdepth}{2}}
\vspace{4mm}
\hrule height .8pt
\vskip 10mm


\section{Benchmark Construction Details}
\label{app:benchmark-details}
\label{app:construction-details}

This appendix section documents the construction-side artifacts behind \benchmark: the sandbox data, tool interface, environment references, traveler profiles and base query and interaction construction.

\subsection{Sandbox Data}
\label{app:sandbox}

\begin{figure}[t]
    \centering
    \includegraphics[width=\columnwidth]{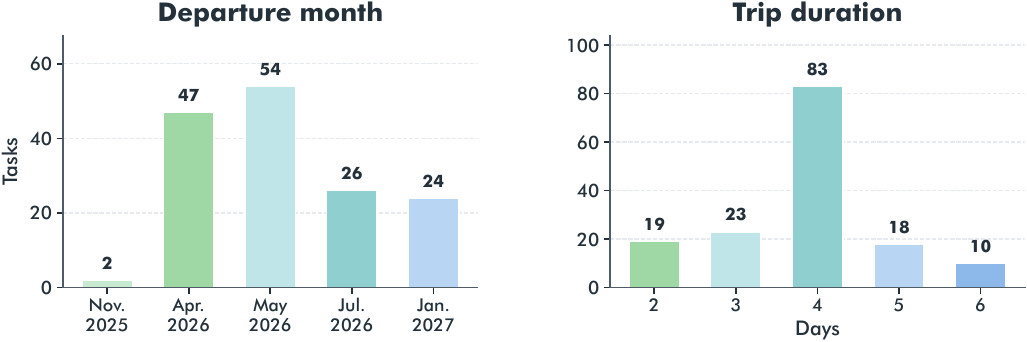}
    \caption{
    Departure-month and trip-duration distributions of benchmark tasks.
    }
    \label{fig:depart}
\end{figure}

The sandbox data provide a fixed, normalized evidence snapshot shared by agents and evaluators. Agents access it through tools, and evaluators use it to verify entity grounding, timing, costs, route continuity, and environment-aware revisions. Price-like fields are reproducible reference costs, not live inventory or fares. Table~\ref{tab:appendix-evidence} summarizes the evidence groups, and Figure~\ref{fig:depart} shows the departure-month and trip-duration distributions of tasks.

\subsection{Tool Interface}
\label{app:tool-interface}
Agents interact with the sandbox through 11 OpenAI-compatible tools. Table~\ref{tab:appendix-tools} groups detailed tools by the evidence they provide.

\subsection{Environment References}
\label{app:environment-references}
The environment layer stores hidden destination-side references, including climate and seasonal context, daily weather, local planning conditions, and environment-event triggers. These references support realistic follow-up turns and let evaluators check whether revised itineraries respond to weather, crowding, traffic, closures, availability changes, and local constraints. Table~\ref{tab:hidden_environmental_context} summarizes the main reference families and their use.

\subsection{Traveler Profile Schema}
\label{app:profile-schema}
Each instance separates traveler information into visible profile cues and hidden evaluation rules. 
For each traveler template, we define candidate observable cues, including party composition, budget range, mobility constraints, pacing preference, interests, and dislikes. 
We sample instance-specific cues and present them to the agent as long-term user memory for itinerary planning. 
The same cues activate hidden profile-specific Rule IDs for soft-preference evaluation, while the Rule IDs and scoring thresholds are never exposed to the agent. 
Table~\ref{tab:profile-derived-soft-rules} maps observable cues to hidden Rule IDs, Table~\ref{tab:soft-rule-score} defines the scoring rubric, and Table~\ref{tab:appendix-profile} summarizes the profile components and their planning/evaluation roles.

\begin{table*}[t!]
\centering
\small
\setlength{\tabcolsep}{6pt}
\renewcommand{\arraystretch}{1.05}
\resizebox{1.0\linewidth}{!}{
\begin{tabular}{L{0.16\textwidth}L{0.46\textwidth}L{0.30\textwidth}}
\toprule
\textbf{Evidence group} & \textbf{Main fields} & \textbf{Benchmark role} \\
\midrule
Attractions &
Name; category; coordinates; rating; opening window; visit duration; ticket cost; popularity and crowd signals &
POI grounding; opening, duration, cost, crowd, and interest checks. \\
\addlinespace

Hotels &
Name; coordinates; star level; price; score; brand; service tags &
Lodging grounding; room count, cost, location, and profile-fit checks. \\
\addlinespace

Restaurants &
Name; coordinates; cuisine; opening window; rating; tags; per-person price &
Meal grounding; cuisine, timing, cost, and local-route checks. \\
\addlinespace

Local entity index &
Canonical names; coordinates; addresses; city anchors; POI type labels &
Location search; entity resolution; endpoint and cross-file grounding. \\
\addlinespace

Local movement &
Origin; destination; distance; duration; local reference cost &
Transfer feasibility; route continuity; local burden, cost, and buffer checks. \\
\addlinespace

Subway evidence &
City file; line; ordered stations; station coordinates when available &
Subway paths; station proximity; transit-mode and rail-access burden checks. \\
\addlinespace

Weather &
City; date; condition; temperature; precipitation-related signals &
Weather-aware construction; outdoor exposure and replanning checks. \\
\addlinespace

Intercity transport &
Endpoints; date; departure/arrival time; duration; class; availability; price &
Flight/train grounding; timing, route, class, availability, and budget checks. \\
\bottomrule
\end{tabular}
}
\caption{Public sandbox evidence exposed through \benchmark tools.}
\label{tab:appendix-evidence}
\end{table*}

\begin{table*}[t!]
\centering
\small
\setlength{\tabcolsep}{6pt}
\renewcommand{\arraystretch}{1.05}
\resizebox{1.0\linewidth}{!}{
\begin{tabular}{L{0.18\textwidth}L{0.29\textwidth}L{0.43\textwidth}}
\toprule
\textbf{Tool family} & \textbf{Tool names} & \textbf{Planning evidence returned} \\
\midrule
Intercity transport &
\texttt{query\_train\_info}; \texttt{query\_flight\_info} &
Candidate routes with endpoints, schedules, duration, class, availability, and prices. \\
\addlinespace
Lodging &
\texttt{query\_hotel\_info} &
Hotels with location, star level, price, score, brand, and services. \\
\addlinespace
Attractions &
\begin{tabular}[t]{@{}l@{}}\texttt{query\_attraction\_details}\\ \texttt{recommend\_attractions}\end{tabular} &
Attractions with location, category, openings, visit duration, ticket cost, rating, popularity, and crowd risk. \\
\addlinespace
Restaurants &
\begin{tabular}[t]{@{}l@{}}\texttt{recommend\_restaurants}\\ \texttt{query\_restaurant\_details}\end{tabular} &
Restaurants with location, cuisine, openings, rating, tags, and per-person price. \\
\addlinespace
Local movement and grounding &
\begin{tabular}[t]{@{}l@{}}\texttt{search\_location}\\ \texttt{query\_road\_route\_info}\\ \texttt{query\_city}\\ \texttt{\_transport\_plan}\end{tabular} &
Entity coordinates; point-to-point distance/duration; route mode; local cost. \\
\addlinespace
Weather &
\texttt{query\_city\_weather} &
Daily city weather for tool grounding and environment-aware planning checks. \\
\bottomrule
\end{tabular}
}
\caption{Tool interface exposed to evaluated agents.}
\label{tab:appendix-tools}
\end{table*}

\paragraph{Rationale for the soft-preference rubric.}
The soft-preference rubric is intended as a transparent and reproducible diagnostic of traveler suitability, rather than as a calibrated psychometric model of tourist satisfaction. Prior tourism research characterizes vacation planning as a contextual, multi-criteria decision process in which travelers differ in constraints, motivations, decision styles, and trade-offs \citep{sirakaya2005building,decrop2005grounded}. We therefore decompose suitability into separate rule families instead of using a single global preference score. The rule families correspond to common sources of travel friction: comfort and pace, local mobility, weather exposure, intercity transport convenience, budget/value, and interest-related coverage. These dimensions are motivated by prior work on tourism fatigue \citep{sun2020development}, weather and climate information for tourism \citep{scott2010weather}, public-transport performance and destination satisfaction \citep{thompson2007investigation}, perceived value in tourism experiences \citep{williams2009value}, and memorable tourism experiences \citep{kim2012development}.

\begin{table*}[t!]
\centering
\small
\setlength{\tabcolsep}{6pt}
\renewcommand{\arraystretch}{1.3}

\begin{tabular}{p{0.22\textwidth} p{0.36\textwidth} p{0.34\textwidth}}
\toprule
\textbf{Context source} & \textbf{Stored reference} & \textbf{Example use} \\
\midrule

Climate and season
& Whether the destination is likely to be hot, cold, rainy, windy, high-altitude, or season-sensitive for the trip month.
& Generates weather-related follow-ups and checks whether the revised plan reduces risky outdoor exposure. \\

Local planning conditions
& Practical city facts that affect planning, such as difficult local transfers, long access routes, or places that need extra buffers.
& Generates local-practical follow-ups and checks whether the plan adds enough transfer or rest buffer. \\

Daily weather
& Weather for each city and date: condition, temperature range, rain amount, rain hours, and location.
& Supports day-level replanning; e.g., move outdoor visits away from rainy periods and add indoor, rest, or buffer blocks. \\

Environment event references
& Event type, affected city/day/time, trigger condition, and required environment-aware response.
& Generates event-grounded follow-ups and checks whether the revised plan adapts to weather, crowding, traffic, closure, or availability changes. \\

\bottomrule
\end{tabular}

\caption{Hidden environment references for turn construction and evaluation.}
\label{tab:hidden_environmental_context}

\end{table*}

\begin{table*}[t!]
\centering
\small
\setlength{\tabcolsep}{6pt}
\renewcommand{\arraystretch}{1.3}
\resizebox{1.0\linewidth}{!}{
\begin{tabular}{L{0.20\textwidth}L{0.18\textwidth}L{0.52\textwidth}}
\toprule
\textbf{Profile component} &\textbf{ Visibility} &\textbf{ Role in evaluation} \\
\midrule
Party composition &
Visible &
Specifies adults, children, and elderly companions; informs room choice, pace, mobility, and safety checks. \\
\addlinespace
Budget range &
Visible &
Captures budget sensitivity; becomes a hard constraint only when the user states an explicit budget bound. \\
\addlinespace
Accommodation style &
Visible &
Guides lodging choice, such as budget, comfort, or luxury preference. \\
\addlinespace
Mobility and physical tolerance &
Visible &
Signals walking limits, stroller or elderly needs, weather sensitivity, and recovery needs. \\
\addlinespace
Schedule rhythm and rest needs &
Visible &
Guides late starts, relaxed pacing, rest blocks, and avoidance of overpacked days. \\
\addlinespace
Interests and dislikes &
Visible &
Guides POI, restaurant, and activity choices; penalizes disliked patterns such as red-eye travel or repeated costly meals. \\
\addlinespace
Transport preferences &
Visible &
Guides choices among trains, flights, taxis, walking, direct routes, transfers, and early departures. \\
\addlinespace
Derived profile rules &
Hidden &
Map visible traits to evaluable checks for walking burden, meal timing, rest, budget pressure, weather exposure, POI fit, and transport burden. \\
\addlinespace
Turn-active profile deltas &
Hidden &
Track profile changes introduced during interaction and determine which expectations remain active after each turn. \\
\bottomrule
\end{tabular}
}
\caption{Traveler profile fields used for planning and evaluation.}
\label{tab:appendix-profile}
\end{table*}

The 1.0/0.5/0.0 scores should be interpreted as coarse ordinal design choices for benchmark evaluation. A score of 1.0 indicates clear alignment with the corresponding traveler preference, 0.5 indicates partial alignment or mild friction, and 0.0 indicates clear mismatch or substantial friction. This middle level is useful because many itinerary-quality issues are graded rather than binary: for example, one long transfer is less severe than several very long transfers, and a slightly dense day is less problematic than an itinerary that repeatedly violates the traveler's pacing preference. Importantly, the cited tourism studies motivate the evaluated dimensions, but do not define our exact numerical thresholds. Future work should calibrate these thresholds and dimension weights with post-trip satisfaction surveys, stated-preference studies, revealed-choice data, human annotation, and sensitivity analyses across traveler groups.

\subsection{Base Query and Interaction Construction}
\label{app:query-construction}

We construct each instance in a state-first manner: the generator first builds a sandbox-grounded base query, then expands it into one of four verifiable multi-turn interaction scenarios. Table~\ref{tab:appendix-construction-pipeline} provides a detailed overview of the construction pipeline and the hidden supervision retained for evaluation.

\begin{table*}[p]
\centering
\setlength{\tabcolsep}{2pt}
\renewcommand{\arraystretch}{0.96}
\scriptsize
\resizebox{1.0\linewidth}{!}{
\begin{tabular}{@{}L{0.10\textwidth}L{0.24\textwidth}L{0.21\textwidth}L{0.21\textwidth}L{0.215\textwidth}@{}}
\toprule
\textbf{Profile} & \textbf{Candidate observable profile cues} & \textbf{Comfort/weather rules} & \textbf{Transport rules} &\textbf{ Budget/interest rules} \\
\midrule
\texttt{P01} Young Backpacker &
moderate/dense pace; heat/cold optional; long local, red-eye, expensive meal; train/flight; food/nature/shopping/landmark &
\texttt{schedule\_pacing}; \texttt{mobility\_accessibility}; \texttt{weather\_avoid\_heat\_exposure}; \texttt{weather\_avoid\_cold\_exposure} &
\texttt{transport\_avoid\_transfer}; \texttt{transport\_avoid\_red\_eye}; \texttt{transport\_prefer\_train}; \texttt{transport\_prefer\_flight} &
\texttt{hotel\_value\_first}; \texttt{budget\_guarded}; \texttt{budget\_tight\_cap}; \texttt{meal\_avoid\_expensive}; \texttt{interest\_local\_food}; \texttt{interest\_outdoor\_nature}; \texttt{interest\_shopping}; \texttt{interest\_landmark} \\
\midrule
\texttt{P02} Honeymoon Couple &
relaxed/moderate pace; cold optional; long local or red-eye; early/flight; food, nature, art, shopping, landmark &
\texttt{schedule\_pacing}; \texttt{mobility\_accessibility}; \texttt{weather\_avoid\_cold\_exposure} &
\texttt{transport\_avoid\_transfer}; \texttt{transport\_avoid\_red\_eye}; \texttt{transport\_avoid\_early\_departure}; \texttt{transport\_prefer\_flight} &
\texttt{budget\_guarded}; \texttt{interest\_local\_food}; \texttt{interest\_outdoor\_nature}; \texttt{interest\_art}; \texttt{interest\_shopping}; \texttt{interest\_landmark} \\
\midrule
\texttt{P03} Three-Generation Family &
relaxed pace/rest; elder/child mobility; heat/cold/extreme weather; late/train; food/nature/culture &
\texttt{schedule\_pacing}; \texttt{mobility\_accessibility}; \texttt{weather\_avoid\_heat\_exposure}; \texttt{weather\_avoid\_cold\_exposure}; \texttt{weather\_need\_backup} &
\texttt{transport\_avoid\_transfer}; \texttt{transport\_avoid\_red\_eye}; \texttt{transport\_avoid\_late\_arrival}; \texttt{transport\_prefer\_train} &
\texttt{budget\_guarded}; \texttt{budget\_tight\_cap}; \texttt{interest\_local\_food}; \texttt{interest\_outdoor\_nature}; \texttt{interest\_culture} \\
\midrule
\texttt{P04} Family with Child &
relaxed/moderate pace; child rest, stroller/walk; heat/extreme weather; late/train; expensive meal; food/nature/culture/shopping &
\texttt{schedule\_pacing}; \texttt{mobility\_accessibility}; \texttt{weather\_avoid\_heat\_exposure}; \texttt{weather\_need\_backup} &
\texttt{transport\_avoid\_transfer}; \texttt{transport\_avoid\_red\_eye}; \texttt{transport\_avoid\_late\_arrival}; \texttt{transport\_prefer\_train} &
\texttt{budget\_guarded}; \texttt{budget\_tight\_cap}; \texttt{meal\_avoid\_expensive}; \texttt{interest\_local\_food}; \texttt{interest\_outdoor\_nature}; \texttt{interest\_culture}; \texttt{interest\_shopping} \\
\midrule
\texttt{P05} Slow-Paced Senior Traveler &
relaxed pace/rest; elder/walk; heat/cold/extreme weather; early/train; food/nature/culture &
\texttt{schedule\_pacing}; \texttt{mobility\_accessibility}; \texttt{weather\_avoid\_heat\_exposure}; \texttt{weather\_avoid\_cold\_exposure}; \texttt{weather\_need\_backup} &
\texttt{transport\_avoid\_transfer}; \texttt{transport\_avoid\_red\_eye}; \texttt{transport\_avoid\_early\_departure}; \texttt{transport\_prefer\_train} &
\texttt{budget\_guarded}; \texttt{budget\_tight\_cap}; \texttt{interest\_local\_food}; \texttt{interest\_outdoor\_nature}; \texttt{interest\_culture} \\
\midrule
\texttt{P06} Cultural Explorer &
moderate pace; cold optional; long local, red-eye, expensive meal; train; food/culture/shopping &
\texttt{schedule\_pacing}; \texttt{mobility\_accessibility}; \texttt{weather\_avoid\_cold\_exposure} &
\texttt{transport\_avoid\_transfer}; \texttt{transport\_avoid\_red\_eye}; \texttt{transport\_prefer\_train} &
\texttt{budget\_guarded}; \texttt{budget\_tight\_cap}; \texttt{meal\_avoid\_expensive}; \texttt{interest\_local\_food}; \texttt{interest\_culture}; \texttt{interest\_shopping} \\
\midrule
\texttt{P07} Budget Student Traveler &
moderate/dense pace; heat optional; long local, expensive meal, red-eye; train/flight; food/nature/culture/shopping &
\texttt{schedule\_pacing}; \texttt{mobility\_accessibility}; \texttt{weather\_avoid\_heat\_exposure} &
\texttt{transport\_avoid\_transfer}; \texttt{transport\_avoid\_red\_eye}; \texttt{transport\_prefer\_train}; \texttt{transport\_prefer\_flight} &
\texttt{hotel\_value\_first}; \texttt{budget\_guarded}; \texttt{budget\_tight\_cap}; \texttt{meal\_avoid\_expensive}; \texttt{interest\_local\_food}; \texttt{interest\_outdoor\_nature}; \texttt{interest\_culture}; \texttt{interest\_shopping} \\
\midrule
\texttt{P08} Business Traveler Extending the Trip &
relaxed/moderate pace; heat optional; long local, red-eye; late/flight; food, culture, shopping, landmark &
\texttt{schedule\_pacing}; \texttt{mobility\_accessibility}; \texttt{weather\_avoid\_heat\_exposure} &
\texttt{transport\_avoid\_transfer}; \texttt{transport\_avoid\_red\_eye}; \texttt{transport\_avoid\_late\_arrival}; \texttt{transport\_prefer\_flight} &
\texttt{budget\_guarded}; \texttt{interest\_local\_food}; \texttt{interest\_culture}; \texttt{interest\_shopping}; \texttt{interest\_landmark} \\
\midrule
\texttt{P09} Food-First Traveler &
relaxed/moderate pace; heat optional; expensive meal, long local, red-eye/late; food-centered &
\texttt{schedule\_pacing}; \texttt{mobility\_accessibility}; \texttt{weather\_avoid\_heat\_exposure} &
\texttt{transport\_avoid\_transfer}; \texttt{transport\_avoid\_red\_eye}; \texttt{transport\_avoid\_late\_arrival} &
\texttt{hotel\_value\_first}; \texttt{budget\_guarded}; \texttt{budget\_tight\_cap}; \texttt{meal\_avoid\_expensive}; \texttt{interest\_local\_food}; \texttt{interest\_outdoor\_nature}; \texttt{interest\_culture}; \texttt{interest\_shopping} \\
\midrule
\texttt{P10} Nature Scenery Lover &
moderate/dense pace; heat/cold/extreme weather; direct/red-eye; nature-centered &
\texttt{schedule\_pacing}; \texttt{mobility\_accessibility}; \texttt{weather\_avoid\_heat\_exposure}; \texttt{weather\_avoid\_cold\_exposure}; \texttt{weather\_need\_backup} &
\texttt{transport\_avoid\_transfer}; \texttt{transport\_avoid\_red\_eye} &
\texttt{hotel\_value\_first}; \texttt{budget\_guarded}; \texttt{budget\_tight\_cap}; \texttt{interest\_local\_food}; \texttt{interest\_outdoor\_nature}; \texttt{interest\_culture} \\
\midrule
\texttt{P11} Friend Group &
moderate/dense pace; heat optional; long local, expensive meal; direct/late; food, nature, shopping, amusement &
\texttt{schedule\_pacing}; \texttt{mobility\_accessibility}; \texttt{weather\_avoid\_heat\_exposure} &
\texttt{transport\_avoid\_transfer}; \texttt{transport\_avoid\_late\_arrival} &
\texttt{hotel\_value\_first}; \texttt{budget\_guarded}; \texttt{budget\_tight\_cap}; \texttt{meal\_avoid\_expensive}; \texttt{interest\_local\_food}; \texttt{interest\_outdoor\_nature}; \texttt{interest\_shopping}; \texttt{interest\_amusement} \\
\bottomrule
\end{tabular}
}
\caption{\textbf{Observable profile evidence and derived evaluator rules.} Each row summarizes possible sampled evidence seeds and the canonical soft-preference rule IDs they can activate. A concrete query uses one sampled subset; unsupported or inapplicable rules are skipped before averaging.}
\label{tab:profile-derived-soft-rules}
\end{table*}
\providecommand{\relprof}{\textcolor{green!45!black}{\textbf{Relax}}}
\providecommand{\modprof}{\textcolor{yellow!45!black}{\textbf{Mod}}}
\providecommand{\denseprof}{\textcolor{red!70!black}{\textbf{Dense}}}

\begin{table*}[t]
\centering
\scriptsize
\setlength{\tabcolsep}{2pt}
\renewcommand{\arraystretch}{2.0}

\vspace{1mm}

\begin{tabular}{@{}L{0.298\textwidth}L{0.214\textwidth}L{0.229\textwidth}L{0.229\textwidth}@{}}
\toprule
\textbf{Rule ID(s) and checked evidence} & \textbf{Score 1.0} & \textbf{Score 0.5} & \textbf{Score 0.0} \\
\midrule

\rowcolor{blue!5}
\multicolumn{4}{@{}L{0.995\textwidth}@{}}{\textbf{Comfort and pace: schedule, local mobility, and weather}} \\

\texttt{schedule\_pacing}: maximum daily attraction count &
\relprof: $\leq$3; \modprof: $\leq$4; \denseprof: $\leq$6 &
\relprof: 4; \modprof: 5--6; \denseprof: 7--8 &
\relprof: $\geq$5; \modprof: $\geq$7; \denseprof: $\geq$9 \\

\texttt{schedule\_pacing}: average daily attraction count &
\relprof: $\leq$3; \modprof: $\leq$4 &
\relprof: $>$3 and $\leq$4; \modprof: $>$4 and $\leq$5 &
\relprof: $>$4; \modprof: $>$5 \\

\texttt{schedule\_pacing}: maximum active day span &
\relprof: $\leq$10 h; \modprof: $\leq$12 h; \denseprof: $\leq$14 h &
\relprof: 10--12 h; \modprof: 12--13.5 h; \denseprof: 14--16 h &
\relprof: $>$12 h; \modprof: $>$13.5 h; \denseprof: $>$16 h \\

\texttt{schedule\_pacing}: earliest daily start &
\relprof: $\geq$08:00 &
\relprof: 07:00--08:00 &
\relprof: $<$07:00 \\

\texttt{schedule\_pacing}: latest daily end &
\relprof/\modprof: $\leq$21:00; \denseprof: $\leq$22:00 &
\relprof/\modprof: 21:00--22:00; \denseprof: 22:00--23:00 &
\relprof/\modprof: $>$22:00; \denseprof: $>$23:00 \\

\texttt{schedule\_pacing}: required rest block &
\relprof: present &
\relprof: missing on light day ($<$5 attractions) &
\relprof: missing on heavy day ($\geq$5 attractions) \\

\texttt{mobility\_accessibility}: maximum within-city transfer time &
$\leq$240 min &
240--300 min &
$>$300 min \\

\texttt{mobility\_accessibility}: long/very-long local transfers &
Route-sensitive: long $<$2 and very-long 0; walking-sensitive: long 0 and very-long 0 &
Route-sensitive: long 2--3 or very-long 1; walking-sensitive: long 1--2 or very-long 1 &
Route-sensitive: long $\geq$4 or very-long $\geq$2; walking-sensitive: long $\geq$3 or very-long $\geq$2 \\

\texttt{mobility\_accessibility}: daily outdoor movement &
$\leq$240 min &
240--360 min &
$>$360 min \\

\texttt{weather\_avoid\_heat\_exposure} / \texttt{weather\_avoid\_cold\_exposure}: outdoor exposure &
Heat: outdoor 11:00--15:00 $\leq$90 min; cold: outdoor after 18:00 $\leq$90 min; total outdoor $\leq$240 min &
Heat/cold period outdoor 90--120 min or total outdoor 240--300 min &
Heat/cold period outdoor $>$120 min or total outdoor $>$300 min \\

\texttt{weather\_need\_backup}: indoor/weather backup &
Backup exists and outdoor load is low ($\leq$240 min) &
Backup missing with non-high outdoor load, or outdoor load 240--300 min &
High outdoor load ($>$300 min), especially without backup \\

\addlinespace[2pt]
\rowcolor{purple!5}
\multicolumn{4}{@{}L{0.995\textwidth}@{}}{\textbf{Transport convenience: intercity transfers, timing, and preferred mode}} \\

\texttt{transport\_avoid\_transfer}: intercity transfer count &
0 transfers &
1 transfer &
$\geq$2 transfers \\

\texttt{transport\_avoid\_red\_eye}: late-night intercity timing &
No severe late-night segment and no late arrival &
1--2 arrivals $>$21:00, and no segment starts $<$06:00 or ends $>$22:00 &
Any segment starts $<$06:00 or ends $>$22:00, or $\geq$3 arrivals $>$21:00 \\

\texttt{transport\_avoid\_early\_departure} / \texttt{transport\_avoid\_late\_arrival}: early/late timing &
No timing issue for the active sensitivity &
Early-sensitive: 1 departure $<$07:00; late-sensitive: 1 arrival $>$21:00 &
Early-sensitive: $\geq$2 departures $<$07:00; late-sensitive: $\geq$2 arrivals $>$21:00 \\

\texttt{transport\_prefer\_train} / \texttt{transport\_prefer\_flight}: preferred mode &
At least one preferred-mode segment &
-- &
No preferred-mode segment \\

\addlinespace[2pt]
\rowcolor{red!5}
\multicolumn{4}{@{}L{0.995\textwidth}@{}}{\textbf{Budget and value: hotel price and meal cost}} \\

\texttt{hotel\_value\_first}: average nightly hotel price &
Avg. nightly hotel $\leq$ city P50 &
City P50--P75, or missing price &
$>$ city P75 \\

\texttt{budget\_guarded} / \texttt{budget\_tight\_cap} / \texttt{meal\_avoid\_expensive}: expensive meal count &
No expensive meals &
Exactly one expensive meal; no very expensive meal &
$\geq$2 expensive meals, or any very expensive meal \\

\addlinespace[2pt]
\rowcolor{green!5}
\multicolumn{4}{@{}L{0.995\textwidth}@{}}{\textbf{Interest match: profile-theme coverage}} \\

\texttt{interest\_local\_food} / \texttt{interest\_outdoor\_nature} / \texttt{interest\_culture} / \texttt{interest\_art} / \texttt{interest\_shopping} / \texttt{interest\_landmark} / \texttt{interest\_amusement}: profile interest theme match &
Profile-theme matches reach target count &
At least one profile-theme match, but below target &
No profile-theme match \\

\bottomrule
\end{tabular}

\caption{\textbf{Rule-ID soft-preference scoring rubric.} Each row shows the canonical evaluator rule ID(s), the itinerary evidence checked by that rule, and the deterministic mapping to 1.0, 0.5, or 0.0. Applicable rule scores are averaged within each dimension, and evaluated dimension scores are then averaged. \textit{Note: P50/P75/P90 are the 50th/75th/90th city price percentiles; required places and explicit must-use modes in the query are hard constraints.}}
\label{tab:soft-rule-score}
\end{table*}

\begin{table*}[t!]
\centering
\small
\resizebox{1.0\linewidth}{!}{
\renewcommand{\arraystretch}{1.5}
\begin{tabular}{L{0.16\textwidth}L{0.10\textwidth}L{0.38\textwidth}L{0.30\textwidth}}
\toprule
\textbf{Pipeline step} & \textbf{Turns} & \textbf{Code-level construction action} &\textbf{ Hidden supervision retained} \\
\midrule
\rowcolor{blue!5}
\multicolumn{4}{@{}l}{\textbf{Base-query construction}} \\
\addlinespace
Trip frame &
T1 source &
\scriptsize
Sample a route option with origin, destination, dates, duration, party size, room count, and intercity mode; build the fixed trip context used by the initial request. &
\scriptsize
Explicit trip constraints such as date range, party size, room count, route mode, and round-trip transport requirements. \\
\addlinespace
Traveler profile &
T1 source &
\scriptsize
Sample one observable traveler profile and derive hidden profile rules from it. The observable profile shapes the user-facing request and planner-visible memory. &
\scriptsize
Hidden profile Rule IDs and scoring bases for soft-preference and traveler-experience checks; Rule IDs and thresholds are not exposed. \\
\addlinespace
Grounded requirements &
T1 source &
\scriptsize
Sample visible database-backed requirements over transport, lodging, restaurants, and attractions; optionally add budget and environment hints when their gates fire. &
\scriptsize
Constraint keys, acceptable candidates, entity evidence, reference costs, soft city/profile wishes, and hidden environment references. \\
\addlinespace
Initial turn wrapper &
T1 &
\scriptsize
Create the first turn with \texttt{sampled\_deltas = [initial\_request]}, \texttt{must\_update = [initial\_plan]}, and the rendered base query as the visible user utterance. &
\scriptsize
Checks that the initial plan satisfies visible hard constraints and uses the observable profile without reading hidden oracle fields. \\
\midrule
\rowcolor{purple!5}
\multicolumn{4}{@{}l}{\textbf{Multi-turn expansion}} \\
\addlinespace
User-state evolution &
4 turns &
\scriptsize
Append three \texttt{\_evolution\_delta} follow-up turns after T1. Each turn samples a distinct user-state family, such as party change, schedule change, added attraction, restaurant requirement, hotel requirement, dietary restriction, or budget update. &
\scriptsize
Each turn stores \texttt{state\_delta}, \texttt{must\_update}, \texttt{must\_preserve}, and checks such as updated party/room counts, added entities, schedule windows, budget caps, or preserved hard constraints. \\
\addlinespace
Request resolution &
3 turns &
\scriptsize
Append two resolution follow-up turns after T1. The first introduces an ambiguity, profile-priority conflict, or conflict with a prior hard constraint; the second either resolves the missing information, relaxes a constraint, or explicitly authorizes no-solution. &
\scriptsize
Expected response mode is part of the oracle: clarification for unresolved ambiguity/conflict, plan after resolution, or no-solution only when the user explicitly asks for an impossibility judgment. \\
\addlinespace
Environment-driven replanning &
3 turns &
\scriptsize
Append two environment follow-up turns after T1 from factors such as weather risk, crowd or queue risk, traffic peak, availability change, transport disruption, or local practical constraint. &
\scriptsize
Each event records factor, event type, references, expected adjustments, and checks that the agent acknowledges the event, revises the plan, and preserves prior hard constraints. \\
\addlinespace
Long-horizon alignment &
5 turns &
\scriptsize
Append four follow-up turns after T1: one user-state evolution turn, followed by resolution, environment, or final-schedule pressure depending on the branch. &
\scriptsize
The oracle accumulates active hard constraints, user-state deltas, request resolutions, and environment events, then checks whether the final response applies all active state while preserving prior commitments. \\
\midrule
 \rowcolor{green!5}
\multicolumn{4}{@{}l}{\textbf{Rendering and validation}} \\
\addlinespace
Turn rendering &
All turns &
\scriptsize
Rewrite each rule-generated turn into a natural user utterance using the previous user turns and the current turn contract. The renderer must preserve \texttt{state\_delta}, \texttt{must\_update}, response-mode cues, and literal anchors. &
\scriptsize
Reject or fall back when the rewrite is empty, too long, loses required anchors, or leaks internal markers such as \texttt{state\_delta}, \texttt{must\_preserve}, \texttt{verification\_oracle}, or oracle fields. \\
\addlinespace
Oracle state update &
All turns &
\scriptsize
After turns are built, accumulate the hidden state with \texttt{\_with\_oracle\_state}: active hard constraints, profile deltas, user-state deltas, request resolutions, environment events, feasibility status, and response expectation. &
\scriptsize
Each turn receives \texttt{oracle\_state\_after\_turn} and \texttt{verification\_oracle}, which define the evaluation target but are never exposed to the planner. \\
\bottomrule
\end{tabular}
}
\caption{State-first construction pipeline for \benchmark~instances. Turn counts start from the initial user request as T1. Base-query construction creates the grounded travel state, while multi-turn expansion adds code-generated state deltas, response-mode expectations, preservation targets, and verification oracles. Only rendered user utterances and observable profiles are exposed to the planner.}
\label{tab:appendix-construction-pipeline}
\end{table*}

\clearpage

\section{Evaluation Metric Details}
\label{app:evaluation-metrics}

This section specifies what each metric reads, what it checks, how it is scored, and which diagnostics are retained for audit.
For itinerary-level evaluation, generated \textsc{Plan} responses are first converted into a structured execution trace, which we call the \textit{converted itinerary}. 
This representation normalizes activities, meals, rest blocks, transfers, lodging, timestamps, durations, costs, and derived itinerary signals, so that plans can be checked deterministically and used for simulation.

\paragraph{Aggregation protocols.}
Unless otherwise noted, model-level scores are obtained by averaging the corresponding per-turn or per-plan scores over the relevant evaluation set. 
We use three aggregation protocols:
\begin{itemize}[leftmargin=*]
    \item \textbf{End-to-end interaction.} 
    \textit{Response-Mode Gating}, \textit{Request Fulfillment}, and \textit{Intent Preservation} are evaluated for every turn-level response and averaged over all evaluated turns.

    \item \textbf{Valid-plan quality.} 
    \textit{Itinerary Feasibility}, \textit{Hard Constraint Satisfaction}, and \textit{Soft Preference Satisfaction} are evaluated only for turns where both the expected and observed response modes are \textsc{Plan}. 
    The reported score is averaged over these eligible generated plans.

    \item \textbf{User simulation.} 
    \textit{User Simulation} is computed on the final generated plan of each task and averaged over tasks.
\end{itemize}

The subsections below define the corresponding per-turn or per-plan scores before aggregation.
\subsection{Response-Mode Gating}
\label{app:response-mode-details}

Response-mode gating is the first step of evaluation. 
The agent prompt allows exactly three final response modes: \textsc{Plan}, \textsc{Clarification}, and \textsc{NoSolution}. 
Let \(m_t\) be the expected response mode and \(\hat{m}_t\) the observed response mode for turn \(t\). 
The turn-level gate score is
\[
R_t=\mathbf{1}[\hat{m}_t=m_t].
\]
If the mode is incorrect, the turn is recorded as a response-mode mismatch and downstream valid-plan quality metrics are skipped. 
Table~\ref{tab:response-mode-gate} summarizes how the response-mode gate defines the applicable conditions, correct response behavior, and representative examples for each response mode.

\begin{table*}[t]
\centering
\small
\setlength{\tabcolsep}{4pt}
\renewcommand{\arraystretch}{1.12}
\resizebox{1.0\linewidth}{!}{
\begin{tabular}{L{0.16\textwidth}L{0.30\textwidth}L{0.27\textwidth}L{0.19\textwidth}}
\toprule
\textbf{Response mode} & \textbf{When the prompt expects it} & \textbf{Correct response behavior} & \textbf{Example case} \\
\midrule
\textsc{Plan} &
Default for complete requests, normal updates, environment changes, resolved priorities, and tool-verifiable revisions. &
Return one complete \texttt{<plan>} with the full updated itinerary: transport, lodging, meals, local movement, timing, and costs. &
The user adds a child, changes the budget, or reports rain; the agent revises the whole itinerary. \\
\addlinespace
\textsc{Clarification} &
Used only for unresolved blocking ambiguity: missing edit target, hard-constraint conflict, or conflict with hard profile facts without a priority decision. &
Return \texttt{<clarification>} asking for the missing decision, priority, or relaxation direction. Do not invent a full plan. &
The user says ``change the restaurant'' without a day, meal slot, location, or candidate set. \\
\addlinespace
\textsc{NoSolution} &
Used only when hard constraints are clear, the user allows a direct impossibility judgment, and tool evidence proves no feasible solution. &
Return \texttt{<no\_solution>} with the grounded blocking reason. Do not include daily plans or reject for soft-preference conflict. &
The user says not to ask follow-up questions, keeps a fixed date and hard budget, and tool prices show every feasible route exceeds that budget. \\
\bottomrule
\end{tabular}
}
\caption{Response-mode gate used before applying itinerary-level metrics.}
\label{tab:response-mode-gate}
\end{table*}

\subsection{Itinerary Feasibility}
\label{app:feasibility-details}

Motivated by verifiable long-horizon travel planning in \citet{zhang2026deepplanning}, 
we define itinerary feasibility as whether a generated trip can be executed in the fixed sandbox. 
The evaluator applies binary atomic checks for structural completeness, sandbox grounding, temporal coherence, venue availability, supported transfers, and cost arithmetic. 
We group these checks into three dimensions: \textit{structure completeness}, \textit{evidence validity}, and \textit{execution operability}. 

Let \(z_c \in \{0,1\}\) denote whether atomic check \(c\) passes, and let \(C_d\) be the set of checks under dimension \(d\). 
The score of each dimension is the average pass rate of its checks:
\[
D_d=\frac{1}{|C_d|}\sum_{c\in C_d} z_c.
\]
The final feasibility score is the average of the three dimension scores:
\[
F_p=\frac{D_{\mathrm{structure}}+D_{\mathrm{evidence}}+D_{\mathrm{operability}}}{3}.
\]
Table~\ref{tab:feasibility-details} lists the atomic diagnostics included in each dimension.
\begin{table*}[t]
\centering
\small
\resizebox{1.0\linewidth}{!}{
\begin{tabular}{L{0.20\textwidth}L{0.31\textwidth}L{0.39\textwidth}}
\toprule
\textbf{Dimension} & \textbf{Subscores averaged inside the dimension} & \textbf{Atomic checks behind the subscores} \\
\midrule
Structure completeness &
Trip duration; route and stay continuity; daily content coverage &
Valid trip duration, closed-loop route structure, seamless intercity transfers, day-boundary continuity, traceable accommodation, hotel-linked day endings, and essential meal or attraction coverage. \\
\addlinespace
Evidence validity &
POI grounding; transport grounding &
Database-grounded accommodations, attractions, meals, intercity transport, and local-move sanity. \\
\addlinespace
Execution operability &
Time and transfer feasibility; venue and duration feasibility; intercity buffer feasibility; budget arithmetic &
Non-overlapping time slots, reasonable transfer time, attraction opening hours, restaurant service hours, closure days, attraction and meal duration, required intercity buffers, and cost calculation correctness. \\
\bottomrule
\end{tabular}
}
\caption{How itinerary feasibility is calculated from atomic checks.}
\label{tab:feasibility-details}
\end{table*}

\subsection{Hard Constraint Satisfaction}
\label{app:hard-constraint-details}

Hard constraints are active user requirements that must be satisfied for the response to answer the task. 
The evaluator dispatches each normalized hard-constraint key to a deterministic checker and computes the plan-level pass ratio:
\[
H_p=\frac{N_{\mathrm{pass},p}}{N_{\mathrm{eval},p}}.
\]
Here \(N_{\mathrm{pass},p}\) and \(N_{\mathrm{eval},p}\) denote the number of passed and evaluated hard checks for plan \(p\). 
Table~\ref{tab:hard-constraint-details} lists each hard-constraint dimension, what is evaluated, and the corresponding calculation target.

\begin{table*}[t]
\centering
\small
\resizebox{1.0\linewidth}{!}{
\begin{tabular}{L{0.24\textwidth}L{0.33\textwidth}L{0.35\textwidth}}
\toprule
\textbf{Hard-constraint dimension} & \textbf{What is evaluated} & \textbf{Calculation target} \\
\midrule
Trip metadata &
Travel dates, number of days, party size, room count, and hard party facts. &
Checks that the itinerary has the required travel dates, number of days, party size, room count, and hard party facts. \\
\addlinespace
Intercity mode and tickets &
Transport mode, route endpoints, date, time window, directness, ticket number, and seat or cabin class. &
Checks required transport mode, route endpoints, date, departure or arrival window, flight or train number, seat or cabin class, directness, and selected candidate set. \\
\addlinespace
Accommodation &
Hotel identity, star level, brand, service, price range, and ranked-choice requirements. &
Checks hotel name, star level, brand, service, price range, cheapest or highest-rated choice, and other visible lodging requirements. \\
\addlinespace
Dining &
Required restaurant, nearby anchor, cuisine, ranking, meal placement, and acceptable candidate set. &
Checks required restaurant, nearby anchor, cuisine or ranking requirement, meal placement, and acceptable restaurant candidate set. \\
\addlinespace
Attractions &
Required or banned attractions, attraction type, ranking, and acceptable candidate set. &
Checks required or banned attractions, attraction type, ranking requirement, and whether selected attractions come from acceptable database-backed candidates. \\
\addlinespace
Budget &
Explicit trip budget cap stated by the user. &
Checks total plan cost against an explicit budget cap when the user states one; profile budget range alone is not treated as a hard budget. \\
\bottomrule
\end{tabular}
}
\caption{Hard-constraint dimensions and deterministic calculation targets.}
\label{tab:hard-constraint-details}
\end{table*}

\subsection{Soft Preference Satisfaction}
\label{app:soft-preference-details}

Soft preferences are profile-derived, implicit expectations that qualitatively shape a high-quality itinerary without acting as rigid hard constraints. 
Their evaluation follows a structured pipeline where the evaluator maps candidate observable profile cues from a traveler template to hidden soft-rule IDs (Table~\ref{tab:profile-derived-soft-rules}) invisible to the agent, inspects itinerary evidence to assign fine-grained scores of 1.0, 0.5, or 0.0 (Table~\ref{tab:soft-rule-score}), and groups active rules into four preference families: \textit{comfort and pace, transport convenience, budget and value,} and \textit{interest match}. 
To prevent families with an asymmetrical, dense number of rules from disproportionately dominating the final metric, we employ a two-step hierarchical aggregation method instead of a naive flat average. 
Specifically, the evaluator first averages the rule scores within each active preference family \(d\):
\[
P_{p,d}=\frac{1}{|R_{p,d}|}\sum_{r\in R_{p,d}}\mathrm{score}(r),
\]
where \(R_{p,d}\) denotes the set of active and applicable rules belonging specifically to preference family \(d\) for plan \(p\). Subsequently, the overall soft preference score \(P_p\) is computed by averaging across all valid families:
\[
P_p=\frac{1}{|A_p|}\sum_{d\in A_p} P_{p,d},
\qquad A_p=\{d: |R_{p,d}|>0\},
\]
where \(A_p\) represents the index set of active preference families that contain at least one applicable rule (i.e., \(|R_{p,d}|>0\)). This family-level aggregation effectively neutralizes rule-density bias, ensuring that all four core preference dimensions contribute equitably to the final evaluation score.

\subsection{Profile-Conditioned User Simulation}
\label{app:user-experience-details}
Following the extraction of deterministic feasibility, hard-requirement, and soft-preference evidence, the profile-conditioned user simulation evaluates the holistic quality of the generated itinerary. 
The simulator ingests a comprehensive context consisting of: 
(i) the \textit{query record} and \textit{turn state} to capture the multi-turn interactive user intent; 
(ii) the \textit{active user profile} defining the traveler's demographic attributes, physical constraints, and budget sensitivities; 
(iii) the \textit{city and environmental contexts} providing localized external constraints such as weather conditions and geographic layouts; 
(iv) the \textit{generated itinerary} representing the planning output under appraisal; and 
(v) the \textit{deterministic activity-level experience trace}. 
This trace serves as the primary factual backbone, providing a fine-grained timeline of the itinerary that chronicles specific time slots for attraction visits, dining, bus transits, and flights, alongside their corresponding cost fields and verification metrics.

The simulator evaluates the itinerary across five distinct experience dimensions detailed in Table~\ref{tab:user-experience-dimensions}: \textit{physical comfort, environmental comfort, schedule comfort, budget comfort,} and \textit{preference satisfaction}. 
Each applicable dimension is scored on a 1--5 rubric according to the explicit qualitative criteria and score anchors defined in Table~\ref{tab:user-simulation-score-anchors}. 
Dimensions without meaningful evidence are dynamically excluded from the evaluation to prevent them from biasing the average toward a default neutral score. 
The final authoritative normalized score \(S \in [0, 1]\) is derived from these dimensions via:
\[
S=\frac{\bar{s}_{\mathrm{dim}}-1}{4},
\]
where \(\bar{s}_{\mathrm{dim}}\) is the mean 1--5 rubric score computed strictly over the applicable experience dimensions.

\begin{table*}[t!]
\centering
\small
\resizebox{1.0\linewidth}{!}{
\begin{tabular}{L{0.20\textwidth}L{0.30\textwidth}L{0.40\textwidth}}
\toprule
\textbf{Experience dimension} & \textbf{What it captures} & \textbf{Example evidence} \\
\midrule
Physical comfort &
Whether the itinerary is physically manageable for the profiled traveler. &
Long walks, long local transfers, high-intensity attractions, early departures, late returns, elders, children, stroller needs, low walking tolerance, high altitude, and missing rest. \\
\addlinespace
Environmental comfort &
Whether destination conditions make the itinerary uncomfortable or risky. &
Heat, cold, rain, altitude, coastal humidity, crowd or queue risk, long outdoor exposure, and whether the plan uses indoor or lower-exposure alternatives. \\
\addlinespace
Schedule comfort &
Whether the traveler has enough time and recovery margin. &
Dense activity sequences, short buffers, tight intercity transfers, delayed meals, excessive waiting, late finishes, and reasonable rest or recovery slots. \\
\addlinespace
Budget comfort &
Whether spending creates stress for this traveler. &
Explicit budget caps, budget-sensitive profiles, total cost relative to the stated budget, unexplained expensive choices, and cost transparency. \\
\addlinespace
Preference satisfaction &
Whether the executed experience matches the user's intended trip. &
Required places, interest-matched attractions or meals, disliked patterns, repetitive choices, comfort upgrades, and whether preferred activities are placed in workable contexts. \\
\bottomrule
\end{tabular}
}
\caption{Five rubric dimensions used to report profile-conditioned user simulation.}
\label{tab:user-experience-dimensions}
\end{table*}

\begin{table*}[t!]
\centering
\small
\setlength{\tabcolsep}{3pt}
\renewcommand{\arraystretch}{1.5}
\resizebox{1.0\linewidth}{!}{
\begin{tabular}{cL{0.17\textwidth}L{0.20\textwidth}L{0.20\textwidth}L{0.20\textwidth}L{0.20\textwidth}}
\toprule
\textbf{Score} & \textbf{Physical comfort} & \textbf{Environmental comfort} & \textbf{Schedule comfort} & \textbf{Budget comfort} & \textbf{Preference satisfaction} \\
\midrule
5 &
Easy movement; ample recovery; fits mobility needs. &
Benign or well-mitigated weather, exposure, altitude, and crowds. &
Relaxed pacing; clear buffers; normal meals and rest. &
Comfortably within budget or clearly justified. &
Strong interest match; dislikes avoided; preferred experiences well placed. \\
\addlinespace
4 &
Mostly comfortable; only minor exertion or transfer burden. &
Minor discomfort, reduced by timing, indoor options, or short exposure. &
Mostly smooth timing with small tradeoffs. &
Reasonable spending with minor pressure. &
Mostly aligned, with small gaps or missed opportunities. \\
\addlinespace
3 &
Acceptable but visibly tiring or effortful. &
Noticeable but tolerable weather, exposure, altitude, or crowd burden. &
Workable schedule with limited buffers or imperfect timing. &
Budget evidence is neutral: neither stressful nor comfortable. &
Acceptable but only partly aligned with preferences. \\
\addlinespace
2 &
Poor fit: long transfers, strenuous activity, or little rest. &
Uncomfortable conditions are not handled well. &
Rushed or fragile timing; tight links or delayed meals. &
Meaningful stress from high cost or tight margin. &
Weak alignment; generic or repeated mismatches. \\
\addlinespace
1 &
Severe burden or clear mobility-profile conflict. &
Severe unmanaged environmental burden or risk. &
Impractical timing; repeated rush or missed recovery. &
Severe stress, over-budget, or unjustified expense. &
Clear conflict with interests, dislikes, or trip style. \\
\bottomrule
\end{tabular}
}
\caption{Score anchors for the 1--5 profile-conditioned user-simulation rubric: A score of 3 indicates relevant but neutral or acceptable evidence; dimensions with no meaningful evidence are marked inapplicable, not defaulted to 3.}
\label{tab:user-simulation-score-anchors}
\end{table*}

\begin{figure*}[t]
\centering
\includegraphics[width=1.0\textwidth]{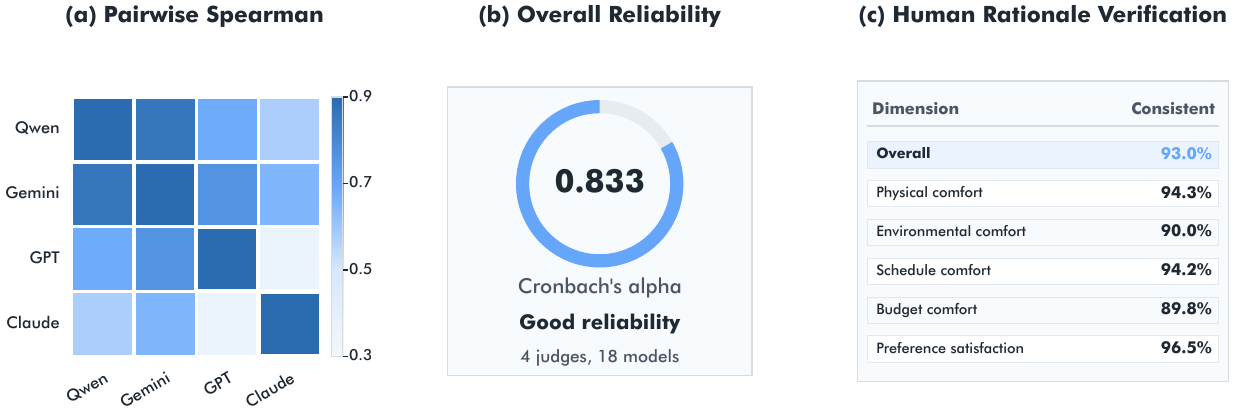}
\vspace{-6mm}
\caption{Reliability and verification of the four user-simulation judges used for median aggregation: (a) pairwise Spearman correlations, (b) overall Cronbach's alpha, and (c) human rationale consistency.}
\label{fig:appendix-user-sim-reliability}
\vspace{-2mm}
\end{figure*}

\subsection{User-Simulation Reliability}
\label{app:user_sim_reliability}


LLM-as-a-judge is widely used for evaluating open-ended generation and instruction-following tasks~\citep{li2025generation, dubois2024length}, but individual judges can show calibration differences and systematic biases~\citep{zheng2023judging}. 
To reduce dependence on a single judge, we use four diverse LLM judges: \textit{qwen3.6-27b}, \textit{gemini-3.1-flash-lite}, \textit{gpt-5.4-nano}, and \textit{claude-haiku-4-5-20251001}. 
Let $s_{i,j}$ denote the normalized user-simulation score assigned to target model $i$ by judge $j$, where $n=18$ target models and $k=4$ judges. 
We report pairwise rank alignment and ensemble consistency over model-level user-simulation scores. 
We further conduct human rationale verification on 50 sampled itineraries, covering 1,825 activity-level evaluations, to check whether scores align with rationales and itinerary segments.

\paragraph{Pairwise Rank Alignment via Spearman's \(\rho\).}
As shown in Figure~\ref{fig:appendix-user-sim-reliability}(a), we evaluate rank agreement between judges using Spearman's rank correlation~\citep{spearman1961proof}:
\[
\rho_{a,b} = 1-\frac{6\sum_{i=1}^{n}d_i^2}{n(n^2-1)},
\]
where \(d_i\) is the rank difference assigned to target model \(i\) by judges \(a\) and \(b\). 
The ensemble yields an average pairwise \(\rho = 0.652\), with the highest agreement observed between Qwen and Gemini (\(\rho = 0.866\)).

\paragraph{Inter-Rater Consistency via Cronbach's \(\alpha\).}
We further report Cronbach's \(\alpha\)~\citep{cronbach1951coefficient} to measure ensemble-level internal consistency:
\[
\alpha = \frac{k}{k-1}
\left(
1 - 
\frac{\sum_{j=1}^{k}\mathrm{Var}_{i=1}^{n}(s_{i,j})}
{\mathrm{Var}_{i=1}^{n}\left(\sum_{j=1}^{k}s_{i,j}\right)}
\right),
\]
where \(\mathrm{Var}_i(\cdot)\) denotes variance over target models. 
The ensemble achieves an overall reliability Cronbach's $\alpha$ of 0.833, as shown in Figure~\ref{fig:appendix-user-sim-reliability}(b).

\paragraph{Human Rationale Verification.}
We audit simulator rationales at the dimension-score level. 
For each sampled itinerary event, we inspect the five score--rationale pairs for \textit{physical comfort}, \textit{environmental comfort}, \textit{schedule comfort}, \textit{budget comfort}, and \textit{preference satisfaction}, and verify whether each pair is consistent with the itinerary segment, traveler profile, cited evidence, and the 1--5 scoring rubric. 
As shown in Figure~\ref{fig:appendix-user-sim-reliability}(c), the results demonstrate a high overall consistency rate of 93.0\%, with preference satisfaction reaching up to 96.5\%, confirming the strong alignment between simulator rationales and human judgments.

The remaining errors are mostly minor boundary or schema issues rather than systematic simulator failures. 
For example, one activity starting at 09:02 was treated as an ``early start'' and assigned a low \textit{schedule comfort} score of 2, although this should only indicate mild schedule pressure under our rubric. 
Some judgments also contain missing or underspecified rationales, and some \textit{environmental comfort} scores fail to consider weather exposure when evaluating outdoor attractions or transfer segments. 
In addition, a few low-cost activities are penalized under \textit{budget comfort} without clear evidence that they create actual budget pressure.

\subsection{Stateful Multi-Turn Evaluation}
\label{app:multiturn-details}

Stateful multi-turn evaluation separates adaptation from memory. At each update turn, request fulfillment measures whether the agent satisfies what the user just added or revised, while intent preservation measures whether it still satisfies earlier commitments that remain active.

Let \(U_t\) be the set of checks introduced or revised at turn \(t\), including hard constraints, soft preferences, and environment conditions:
\[
U_t=\Delta_t^{H}\cup\Delta_t^{Q}\cup\Delta_t^{E}.
\]
The turn-level request-fulfillment score is the pass rate over this current-update set:
\[
F_t=\frac{\sum_{u\in U_t}\mathbf{1}[\mathrm{pass}(u,t)]}{|U_t|}.
\]

Let \(P_t\) be the set of earlier hard constraints, soft preferences, and environment conditions that remain active after turn \(t\). Commitments explicitly removed or revised by the user are excluded. Intent preservation is the pass rate over this preserved set:
\[
I_t=\frac{\sum_{p\in P_t}\mathbf{1}[\mathrm{pass}(p,t)]}{|P_t|}.
\]
Final scores average only over turns with non-empty denominators:
\[
\overline{F}=\frac{1}{|\mathcal{T}_{F}|}\sum_{t\in\mathcal{T}_{F}}F_t,
\qquad
\overline{I}=\frac{1}{|\mathcal{T}_{I}|}\sum_{t\in\mathcal{T}_{I}}I_t.
\]
Here \(\mathcal{T}_{F}\) contains turns with at least one current-update check, and \(\mathcal{T}_{I}\) contains turns with at least one preserved prior commitment. Turn~1 has no prior commitments, so it is excluded from intent preservation. Reporting both scores prevents a model from scoring well by only following the newest update or only preserving the old plan.

\section{Experiments Details}
\label{app:exp-detail}

\begin{figure*}[t]
\centering
\includegraphics[width=\textwidth]{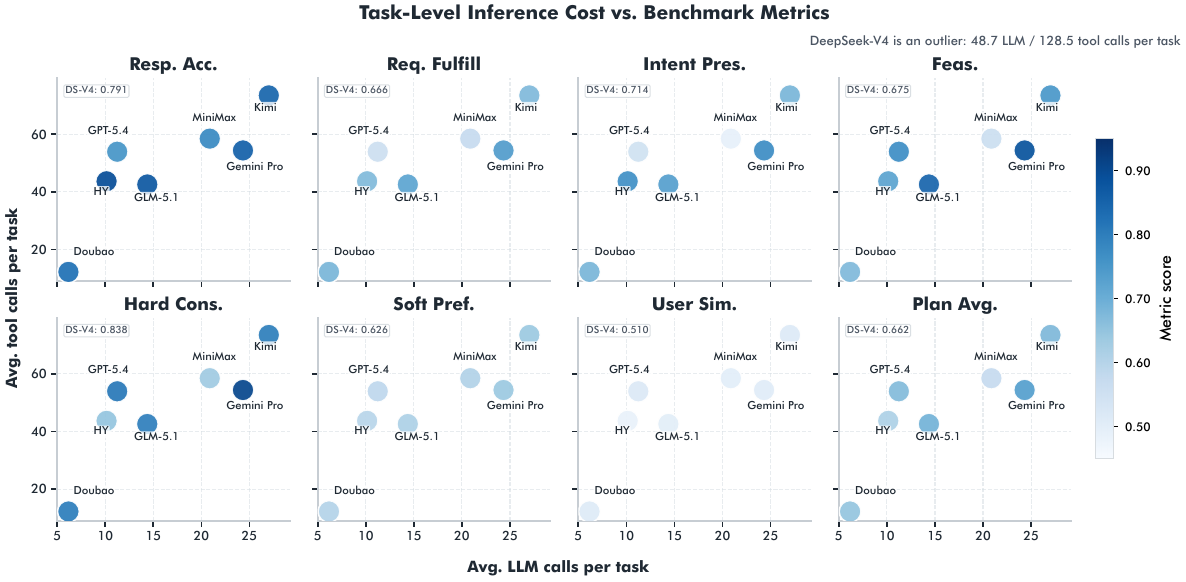}
\vspace{-6mm}
\caption{
Task-level inference cost and planning performance across benchmark metrics
}
\vspace{-2mm}
\label{fig:cost-all}
\end{figure*}

\subsection{Model Set} \label{app:model}

We evaluate 18 tool-using LLM agents. The API-served group includes
Gemini-3.1-Pro-Preview~\citep{google2026gemini31pro},
Gemini-3-Flash-Preview~\citep{google2025gemini3flash},
GPT-5.4~\citep{openai2026gpt54},
GPT-5.4-Mini~\citep{openai2026gpt54mini},
GLM-5.1~\citep{zai2026glm51},
Kimi-K2.6~\citep{moonshot2026kimik26},
Doubao-Seed-2.0-Pro~\citep{bytedance2026seed20},
DeepSeek-V4-Pro~\citep{deepseek2026v4preview},
DeepSeek-V3.2~\citep{liu2025deepseek},
MiniMax-M2.7~\citep{minimax2026m27},
and Hy3-preview~\citep{tencent2026hy3preview}.
All API-served models are accessed through OpenAI-compatible remote endpoints
and exposed to the same function-calling interface.

The lightweight/open-weight group includes Gemma-4-31B, Gemma-4-26B-A4B, Qwen3.5-27B, Qwen3.5-122B-A10B-FP8, Qwen3.6-27B, Qwen3.6-35B-A3B, and GLM-4-32B~\citep{google2026gemma4,qwen2026qwen35,qwen2026qwen3627b,qwen2026qwen3635ba3b,zai2025glm432b0414}. These models are served locally with vLLM~\citep{kwon2023efficient} on 8 NVIDIA A100 80GB GPUs.

\subsection{Model Parameters and Serving} \label{app:model_p}

All models are evaluated with deterministic generation whenever the endpoint supports it. The default settings are temperature 0.0, top-\(p=1.0\), and a maximum of 100 LLM calls per turn.

All target models are exposed to the benchmark through the same OpenAI-compatible function-calling scaffold. Local vLLM services use tensor parallelism with two to four A100 PCIe 80GB GPUs per model, bfloat16 inference, a 32,768-token maximum context unless explicitly overridden, prefix caching, chunked prefill, up to 32 concurrent sequences, and a maximum batched-token budget of 65,536. For local vLLM models, model-family-specific parsers and chat templates are used only inside the serving layer to translate model-native outputs into this common interface: Qwen uses the Qwen tool-call parser with thinking disabled, while Gemma and GLM use Gemma- and GLM-compatible chat templates and tool-call parsers.

\subsection{Inference Cost Analysis} \label{app:inference}
We further examine whether higher inference effort leads to better planning quality. We use average LLM and tool calls per task as lightweight cost proxies, since each task requires evidence retrieval, itinerary construction or revision, and cross-turn state tracking. Figure~\ref{fig:cost-all} shows no clear monotonic relationship between inference cost and benchmark performance across response-mode, constraint-satisfaction, feasibility, preference, user-simulation, and overall planning metrics.

Higher-cost models do not necessarily produce better plans: DeepSeek-V4 is a clear outlier with 48.7 LLM calls and 128.5 tool calls per task, yet it is not consistently the best model across metrics (e.g., dropping to 0.510 on User Sim.). Kimi also uses substantially more inference than most other frontier models, but its gains are uneven: it performs strongly on several rule-based metrics but does not dominate user-simulation quality.

By contrast, Gemini Pro achieves the strongest overall planning quality without reaching the extreme inference costs of the outliers. Furthermore, models like GPT-5.4 and GLM-5.1 demonstrate highly efficient and competitive performance using roughly half the LLM calls of Gemini Pro and Kimi, while Doubao remains both low-cost and weak. This suggests that \textit{effective evidence use and reasoning matters more than simply increasing the number of calls}.
\section{Prompt Templates}
\label{appendix_prompts}

This appendix lists the agent framework prompt and the profile-conditioned user-simulation prompt used by the implementation. Each blue box inlines the corresponding prompt text.

\definecolor{PromptBlueBg}{HTML}{F3FAFF}
\definecolor{PromptBlueTitle}{HTML}{DCEEFF}
\definecolor{PromptBlueFrame}{HTML}{8CBCE8}

\vspace{4mm}
\tcbset{
    promptbox/.style={
        title={#1},
        colback=PromptBlueBg,
        colframe=PromptBlueFrame,
        colbacktitle=PromptBlueTitle,
        coltitle=black,
        fonttitle=\bfseries,
        boxrule=0.5pt
    }
}

\lstdefinestyle{promptlisting}{
    basicstyle=\ttfamily\footnotesize,
    breaklines=true,
    breakatwhitespace=false,
    columns=fullflexible,
    showstringspaces=false,
    keepspaces=true,
    upquote=true,
    literate={¥}{{\textyen}}1 {￥}{{\textyen}}1
}

\begin{tcolorbox}[
    promptbox={Agent Framework Prompt},
    width=\linewidth,
    breakable,
    enhanced,
    boxsep=2pt,
    left=4pt,
    right=4pt,
    top=4pt,
    bottom=4pt,
    before skip=4pt,
    after skip=6pt
]
\footnotesize
\begin{lstlisting}[style=promptlisting]
You are a travel-planning assistant. Use the current user request, the visible traveler profile, and tool results to return an executable, verifiable, parseable travel response.

The final answer must use exactly one mode: a complete `<plan>`, a complete `<clarification>`, or a complete `<no_solution>`. Do not output tool-call XML, analysis, drafts, recalculation notes, self-corrections, multiple budgets, or a plan followed by another conclusion.

================================================================
Workflow
================================================================
1. Decide the response mode first: plan by default; use `<clarification>` or `<no_solution>` only for unresolved blocking cases listed under Response Modes.
2. Collect the required tool evidence: intercity transport, accommodation, required or candidate attractions, required restaurants, and the adjacent intracity transfers that will actually appear in the final itinerary.
3. Return the complete final result: once evidence is sufficient, produce one complete `<plan>`; if tool results rule out an early candidate, update the candidate and output only the final version, not the trial-and-error process.

================================================================
Core Contract
================================================================
- Satisfy explicit user hard constraints first. The visible profile is for personalization unless it contains hard facts such as party size, companions, mobility, or safety limits.
- In multi-turn dialogue, previously confirmed hard constraints, profile preferences, environment events, and user tradeoffs remain active unless the user explicitly cancels, replaces, or relaxes them. The latest turn does not reset the task.
- Entities, times, prices, routes, and transport facts in the plan must come from tool results. Do not fill gaps from common knowledge. Query user-named places exactly first.
- All locatable entity names must exactly match tool-returned names, including hotels, attractions, restaurants, transport hubs, and `travel_city` endpoints. Entity-name fields should contain only the name, not explanatory text. Put activity purpose or status in the `buffer` description, the `hotel` action, or the scheduled time instead.
- If the user asks to modify an earlier itinerary, still output the updated complete `<plan>`, not only the edited fragment.

================================================================
Response Modes
================================================================
Plan directly by default. Use `<clarification>` only for these unresolved blocking cases:
- The latest message is only a local add/change request for an itinerary component, such as dining, attraction, lodging, or transport, but gives no usable date, time slot, itinerary place, transport node, or candidate set, so you cannot tell which part should change.
- The latest message introduces a new hard constraint that conflicts with still-active previous hard constraints, and the user has not stated the priority or relaxation direction.
- The latest message introduces a new hard request that would clearly override hard facts or strong preferences in the visible profile, such as party composition, children/elders, mobility, safety limits, relaxed pacing, reduced walking, or recovery needs, and the user has not said whether the new goal or profile comfort has priority.

Return a complete `<plan>` for complete initial travel requests, normal user-state updates, normal environment changes, already prioritized or relaxed requests, and any itinerary revision that can be verified and executed with tools. When returning a plan, preserve all still-active constraints and preferences from earlier turns. Decide restaurant anchors, attraction order, hotel area, transport tradeoffs, and rating/price/opening-hour tie-breaks from tool results; do not ask about these execution details.
For complete initial requests, do not output `<clarification>` for these execution details: whether the stated room count is really needed, choosing among hotel/apartment/homestay candidates, choosing highest-rated or cheapest candidates, deciding whether a restaurant can be lunch or dinner from business hours, selecting a return train/flight time, choosing an imperfect but budget-compatible hotel area, or deriving hotel nights from explicit dates. If a tool-returned executable candidate exists, choose the candidate that best satisfies the explicit constraints and output the plan.

Use `<no_solution>` only when all of the following are true:
- The current active hard constraints are clear; no date, place, candidate set, priority, or relaxation direction is still missing.
- The user has authorized a direct no-solution judgment, such as saying not to ask follow-up questions, not to silently replace requirements, or not to relax earlier constraints.
- Tool evidence or known constraints prove that the active hard constraints cannot all be satisfied, such as no available transport on the required date, an explicit budget below the minimum verifiable cost, a required entity unavailable on the usable date, or a new condition that the user insists on keeping despite conflict with an active hard constraint.

If the missing piece is required information, priority, or relaxation direction, output `<clarification>`. If the user has not authorized a direct no-solution judgment, ask which constraint can be relaxed instead of outputting `<no_solution>`.
Do not output `<no_solution>` because of soft-preference conflict, few candidates, imperfect experience, missing opening-hour fields, or facts that should be checked with tools. Do not fabricate entities, prices, routes, business hours, or transport schedules to avoid `<clarification>` or `<no_solution>`.

Entity existence, opening hours, prices, distances, and routes are tool-verification duties, not clarification reasons. An explicit user budget is a hard constraint; profile `budget_range` is not a hard budget for this trip.

================================================================
Tool Evidence Requirements
================================================================
- Intercity transport: use `query_flight_info` or `query_train_info`; pass city names as `origin/destination`. The returned `price` is the complete per-person reference price for the candidate route. For connecting routes, write each segment with its own number, stations, and times; count the same-day route-level `price` only once in the budget.
- Accommodation: for overnight trips or room requests, use `query_hotel_info`; for a named hotel/apartment/homestay, use exact `hotelName`. Planned hotel names and prices must come from the tool.
- Attractions: before scheduling an attraction, use `query_attraction_details` for opening information, duration, and ticket price. Do not schedule attractions that the tool clearly marks as closed.
- Restaurants: before scheduling a restaurant, use `recommend_restaurants` or `query_restaurant_details`. For a named restaurant, use details lookup; for eating near a place, use that place only as the restaurant-search anchor.
- Every `meal` line must contain a specific tool-returned restaurant name and per-person price. Across the complete itinerary, prefer a different restaurant for each meal, and every restaurant name must come from tool results, unless the user explicitly asks to revisit one; one restaurant should normally be used for only one meal. Avoid repeating the same restaurant when possible, and do not use a generic restaurant name. Do not write breakfast, self-arranged light meals, or rest as ungrounded `meal` entities; use a `buffer` or `hotel` description when needed.
- Intracity transport evidence: when two adjacent activities in the same city happen at different places, insert a `travel_city` segment between them and verify its route, duration, distance, and price with tools. Default to `query_city_transport_plan`, especially for named places or when the user prefers metro/subway, fewer transfers, less walking, lower cost, or shorter travel time; keep the returned mode/line summary in the final itinerary. `query_road_route_info` is only a coordinate-level fallback: use `search_location` + `query_road_route_info` only when you already have two exact coordinates and do not need metro/subway line planning.
- Pass only necessary tool arguments: dates/times are used only for flights, trains, and weather; do not pass dates/times to hotels, intracity transport, road routes, attractions, restaurants, or location search. If a tool call repeats identical normalized arguments, reuse the previous result and continue with necessary new arguments.
- Weather: for date-specific trips, you may use `query_city_weather`; if tools show clear weather risk, reflect a reasonable adjustment in the plan.
- Comparative requirements such as cheapest, highest-rated, closest, required cabin/seat class, or time windows are hard constraints and should be judged within the corresponding tool-returned candidate set.

================================================================
Planning Rules
================================================================
- Provide a minute-level timeline for actual travel-related activities.
- Time and location must be continuous: use `travel_city` between different places, `travel_intercity_public` for intercity segments, and `buffer` for procedures, waits, or short rests.
- Activity times must be compatible with transport schedules, opening/business hours, route order, and the visible profile.
- Meal rules: do not schedule breakfast; assume it is handled at the hotel or before departure, and do not count breakfast toward required meals. On a full sightseeing day in the destination city, schedule lunch and dinner. On intercity days, decide meals from the effective time in the destination city: if arriving before 10:00, schedule lunch and dinner; if arriving 10:00-15:00, schedule dinner and lunch is optional; if arriving after 15:00, schedule no meal or only dinner. If departing the destination before 09:00, schedule no meal in that city; if departing 09:00-15:00, lunch is optional and dinner should not be scheduled; if departing after 15:00, schedule at least lunch and dinner is optional.
- Meal timing: lunch should preferably fit within 11:00-14:00, and dinner should preferably fit within 17:00-20:00. Each meal usually takes 1-2 hours. If a day has both lunch and dinner, lunch end and dinner start should be at least 3 hours apart. Restaurant business hours must cover the corresponding meal slot.
- Non-final days should end at that night's accommodation. Final day uses `Accommodation: -`.
- Before flights, write a 90-minute airport buffer; after flight arrival, write at least 30 minutes before leaving the airport. Before trains, write a 30-minute station buffer; after train arrival, write at least 15 minutes before leaving the station.
- `travel_city` duration should be close to the tool-returned duration. `travel_intercity_public` times must match tool results. Attraction duration must fall within the tool-returned range.

================================================================
Output Format
================================================================
<plan>
Day [Day Number] ([YYYY-MM-DD]):
Current City: [from origin city to destination city / city name]
Accommodation: [tool-returned hotel name, ¥positive/room/night; use - on final day]
HH:MM-HH:MM | buffer | [security/waiting; deplaning/exiting; baggage claim; necessary short wait]
HH:MM-HH:MM | travel_intercity_public | [flight/train] [returned number], [returned departure station] - [returned arrival station], [cabin/seat class], ¥[positive]/person
HH:MM-HH:MM | travel_city | [from] - [to], [taxi/walking/metro lines], [distance], [duration], ¥[price]
HH:MM-HH:MM | attraction | [returned attraction name], ¥[ticket]/person
HH:MM-HH:MM | meal | [lunch/dinner], [returned restaurant name], ¥[per person]/person
HH:MM-HH:MM | hotel | [check-in/check-out/rest], [returned hotel name]

**Budget Summary**:
**Transportation: X RMB**. Intercity tickets = one route-level price per same-day flight/train connection * people. Intracity transport depends on the tool-returned mode: taxi/cab prices are per vehicle/trip and should be multiplied by required vehicles (default taxi capacity: 4 people, rounded up); metro/bus/public-transit prices are per person; walking costs 0.
**Accommodation: X RMB**. Hotel price * room count * nights.
**Meals: X RMB**. Per-person meal prices * people.
**Attractions & Tickets: X RMB**. Ticket prices * people.
**Other: X RMB**
**Total Estimated Budget: X RMB**
</plan>

<clarification>
[Ask one or two short questions naming the missing slot, conflicting constraints, or priority that the user must confirm.]
</clarification>

<no_solution>
The current hard constraints cannot be jointly satisfied.
Blocking constraints: name the mutually conflicting or impossible user hard constraints.
Tool evidence: cite the key tool results, such as transport, accommodation, required meal, or ticket costs.
To continue planning, the user would need to relax: list at most two relaxation directions.
</no_solution>
\end{lstlisting}
\end{tcolorbox}

\begin{tcolorbox}[
    promptbox={Profile-Conditioned User-Simulation Prompt},
    width=\linewidth,
    breakable,
    enhanced,
    boxsep=2pt,
    left=4pt,
    right=4pt,
    top=4pt,
    bottom=4pt,
    before skip=4pt,
    after skip=6pt
]
\footnotesize
\begin{lstlisting}[style=promptlisting]
# Profile-Conditioned User Experience

You are the traveler described by `EXPERIENCE_TRACE.user_model`. You are now experiencing the itinerary in `PLAN`, one activity at a time, under the environment and budget conditions in `EXPERIENCE_TRACE.environment` and `EXPERIENCE_TRACE.budget`.

Describe your activity-level travel experience and score the itinerary's experience quality. Your scores must follow the 1-5 rubric below.

Base the experience report on `EXPERIENCE_TRACE`. Use `PLAN` only as a raw itinerary reference for the matching activity position, such as activity name, type, time slot, cost, transport mode, and accommodation. For example, `D1-A1` maps to `PLAN.daily_plans[0].activities[0]`. Do not use `PLAN` to re-judge entity grounding, factual validity, route validity, prices, hard constraints, or requirement success.

`EXPERIENCE_TRACE.activity_trace[].experience_facts` contains neutral facts, not pre-scored verdicts. Infer comfort scores from those facts and the rubric. Boolean conditions are represented only as positive `experience_flags`; absence of a flag means the condition is not evidenced. Do not assume that every cost is budget stress. Treat minor costs as neutral for that activity; use activity cost for budget discomfort only when `budget_cost_relevance` is moderate/high. Use near-limit/over-limit budget margin as whole-trip budget context, not as a reason to penalize every paid activity equally.

Return JSON only. No markdown fences, comments, or extra text.

## Core Rule

Output exactly one `activity_simulations[]` item for each `EXPERIENCE_TRACE.activity_trace[]` item.

Preserve every expected ref exactly once:

```text
EXPERIENCE_TRACE.expected_activity_refs == activity_simulations[].item_ref
```

If evidence is weak or missing, keep the claim neutral or low-confidence and list the missing evidence in `missing_evidence`.

## Evidence Priority

Use evidence in this order. Later sources may clarify raw fields, but they must not override earlier travel-experience evidence.

1. `EXPERIENCE_TRACE.activity_trace[]`: one item per planned activity. Each item contains `item_ref`, compact `event` fields, and neutral `experience_facts` such as duration bucket, positive experience flags, and cost relevance. Use this as the primary source for every `activity_simulations[]` item.
2. `EXPERIENCE_TRACE.user_model`: the compact traveler profile. It may contain `party`, `comfort_sensitivities`, `interest_preferences`, and `sensitivity_flags`. Use it only to interpret how this traveler would feel.
3. `EXPERIENCE_TRACE.environment`: destination or trip-context environment signals, such as heat, cold, rain, exposure, or other risk tags. Use it for environment-related comfort only when connected to an activity.
4. `EXPERIENCE_TRACE.budget`: budget applicability, budget limit when available, estimated total, and margin level. Use it for `budget_comfort`; set budget dimensions not applicable when budget is not applicable.
5. `PLAN`: compact raw itinerary fields, including `daily_plans[]`, `activities[]`, `accommodation`, and `budget_summary`. Use it only to confirm raw fields for the matching activity position, where `D{day}-A{activity_index}` maps to `PLAN.daily_plans[day-1].activities[activity_index-1]`. Do not use it to invent subjective quality, factual validity, route validity, or hidden preferences.

Do not invent facts that are not in the inputs. For example, do not assume extra rest, sleep quality, delays, crowds, weather, closures, scenic quality, restaurant quality, health conditions, preferences, or budget sensitivity.

## Scores

Use a 1-5 scale where higher is better:

- 5: very good, low burden and strongly profile-aligned
- 4: good, minor issues
- 3: acceptable with visible tradeoffs
- 2: poor, uncomfortable or stressful
- 1: very poor, severe discomfort, stress, or profile conflict

Return these five dimensions:

- `physical_comfort`: walking, transfers, standing, stamina, recovery
- `environmental_comfort`: weather, temperature, rain/snow, exposure, crowds when evidenced
- `schedule_comfort`: early starts, late finishes, tightness, buffers, meal timing, density
- `budget_comfort`: budget pressure from relevant costs or tight budget margin; set `applicable=false` if no budget cap/sensitivity exists
- `preference_satisfaction`: interests, dislikes, pace, dining, hotel, transport preferences

Use the same dimensions inside every activity's `dimension_updates` and in top-level `experience_dimensions`.
For each activity, include all five dimensions in `dimension_updates`, but set `applicable=false` and `score_1_5=null` when that activity has no direct evidence for a dimension. Do not fill unrelated dimensions with 3 just to avoid null. Use score 3 only when a dimension is genuinely relevant and the evidenced experience is neutral or acceptable.
For top-level `experience_dimensions`, give the traveler's standardized five-dimension judgment for the whole itinerary or current chunk. Mark a top-level dimension `applicable=false` when there is no meaningful evidence for that dimension anywhere in the chunk. Cite the key `item_ref` values behind each applicable dimension.

`llm_reported_overall.dimension_analysis` explains the holistic overall score and may weight activities by trip importance. `experience_dimensions` are the standardized dimension scores used by the evaluator. Keep both grounded in cited `item_ref` values.

## Activity Fields

For each activity simulation, include:

- `item_ref`, `day`, `activity_index`
- compact `activity`: `type`, `name`, `time_slot`
- `dimension_updates` for all five dimensions
- one or two grounded `evidence` entries
- `confidence`: `high`, `medium`, or `low`

For each evidence entry, `source` names where the support came from, and `claim` states what that source supports. Prefer sources from `EXPERIENCE_TRACE`; use `PLAN` only for activity identity, time, and cost verification. Valid source formats include `EXPERIENCE_TRACE.activity_trace[D1-A1]`, `EXPERIENCE_TRACE.user_model`, `EXPERIENCE_TRACE.environment`, `EXPERIENCE_TRACE.budget`, and `plan.daily_plans[0].activities[0]`.

Confidence:

- `high`: all important claims are directly supported by trace evidence
- `medium`: evidence-backed but partly indirect
- `low`: missing, weak, vague, or inferred evidence affects the score

Use the lowest confidence triggered by any important activity claim. Do not mark every item `high`.

## Required JSON Shape

Keep free-text fields short: one sentence, preferably under 12 words.

Top-level keys:

```text
llm_reported_overall
profile_summary
activity_simulations
experience_dimensions
missing_evidence
audit_notes
```

Minimal shape:

```json
{
  "llm_reported_overall": {
    "score_1_5": 3.0,
    "reason": "one short evidence-based explanation",
    "dimension_analysis": {
      "physical_comfort": {"score_1_5": 3.0, "reason": "", "evidence": ["D1-A1"]},
      "environmental_comfort": {"score_1_5": 3.0, "reason": "", "evidence": ["D1-A1"]},
      "schedule_comfort": {"score_1_5": 3.0, "reason": "", "evidence": ["D1-A1"]},
      "budget_comfort": {"score_1_5": null, "applicable": false, "not_applicable_reason": ""},
      "preference_satisfaction": {"score_1_5": 3.0, "reason": "", "evidence": ["D1-A1"]}
    },
    "authoritative": false
  },
  "profile_summary": {
    "party": {},
    "comfort_sensitivities": {},
    "interest_preferences": {},
    "sensitivity_flags": {},
    "profile_uncertainties": []
  },
  "activity_simulations": [
    {
      "item_ref": "D1-A1",
      "day": 1,
      "activity_index": 1,
      "activity": {"type": "", "name": "", "time_slot": ""},
      "dimension_updates": {
        "physical_comfort": {"score_1_5": 3.0, "applicable": true, "reason": ""},
        "environmental_comfort": {"score_1_5": null, "applicable": false, "not_applicable_reason": ""},
        "schedule_comfort": {"score_1_5": 3.0, "applicable": true, "reason": ""},
        "budget_comfort": {"score_1_5": null, "applicable": false, "not_applicable_reason": ""},
        "preference_satisfaction": {"score_1_5": null, "applicable": false, "not_applicable_reason": ""}
      },
      "evidence": [
        {"item_ref": "D1-A1", "source": "EXPERIENCE_TRACE.activity_trace[D1-A1]", "claim": "", "score_impact": "neutral"}
      ],
      "confidence": "medium"
    }
  ],
  "experience_dimensions": {
    "physical_comfort": {"score_1_5": 3.0, "applicable": true, "evidence": ["D1-A1"]},
    "environmental_comfort": {"score_1_5": 3.0, "applicable": true, "evidence": ["D1-A1"]},
    "schedule_comfort": {"score_1_5": 3.0, "applicable": true, "evidence": ["D1-A1"]},
    "budget_comfort": {"score_1_5": null, "applicable": false, "not_applicable_reason": ""},
    "preference_satisfaction": {"score_1_5": 3.0, "applicable": true, "evidence": ["D1-A1"]}
  },
  "missing_evidence": [],
  "audit_notes": []
}
```

The evaluator will compute normalized scores from `score_1_5`; do not add extra scoring fields.

## Final Checks

Before returning JSON, verify:

1. Every expected activity ref appears exactly once.
2. Every activity has non-empty evidence.
3. Every evidence entry names both the supporting source and the supported claim.
4. Every applicable score is between 1 and 5.
5. No claim depends on invented facts.
\end{lstlisting}
\end{tcolorbox}


\section{Query and Plan Field Examples}
\label{appendix_prompts}

This appendix presents one representative query record and its corresponding plan record to illustrate the released data format. 
The query record includes the user-visible multi-turn request, active constraints, requested update, and response contract used to specify the expected planning behavior. 
The plan record is a lightly normalized model output from our evaluation pipeline, included to demonstrate the required minute-level itinerary format. 
We remove raw tool traces, evaluator-side annotations, run identifiers, and implementation metadata that are not needed to interpret the example.

\definecolor{PromptBlueBg}{HTML}{F3FAFF}
\definecolor{PromptBlueTitle}{HTML}{DCEEFF}
\definecolor{PromptBlueFrame}{HTML}{8CBCE8}
\vspace{4mm}
\tcbset{
    promptbox/.style={
        title={#1},
        colback=PromptBlueBg,
        colframe=PromptBlueFrame,
        colbacktitle=PromptBlueTitle,
        coltitle=black,
        fonttitle=\bfseries,
        boxrule=0.5pt
    }
}

\lstdefinestyle{promptlisting}{
    basicstyle=\ttfamily\footnotesize,
    breaklines=true,
    breakatwhitespace=false,
    columns=fullflexible,
    showstringspaces=false,
    keepspaces=true,
    upquote=true,
    literate={¥}{{\textyen}}1 {￥}{{\textyen}}1
}

\begin{tcolorbox}[
    promptbox={Query Field Example (JSON)},
    width=\linewidth,
    breakable,
    enhanced,
    boxsep=2pt,
    left=4pt,
    right=4pt,
    top=4pt,
    bottom=4pt,
    before skip=4pt,
    after skip=6pt
]
\footnotesize
\begin{lstlisting}[style=promptlisting]
{
  "id": "mt_single_0005_turn_1",
  "source_item_id": "mt_single_0005",
  "query": {
    "base_query_id": "single_0005",
    "interaction_type": "long_horizon_alignment",
    "interaction_label": "Long-Horizon Alignment",
    "turn_count_in_dataset": 5,
    "target_turn_id": 1,
    "response_expectation": "plan",
    "resolution_need": "none",
    "feasibility_status": "solved",
    "visible_turn_history": [
      {
        "turn_id": 0,
        "role": "user",
        "utterance": "My partner and I are planning a nature-focused getaway from Beijing to Shenzhen, arriving on July 14, 2026 (Tuesday) and staying for 4 days, with our return flight on July 17, 2026 (Friday). We'll need just one room. Since we love scenic parks, our main goal is to visit Nantou Ancient City. To keep things easy, we'd like to book a single meal near \"Nantou Ancient City Museum\". For accommodation, we're looking for the highest-rated hotel in the city, but we need to stick to a budget between 5400 and 7200 yuan.",
        "sampled_deltas": ["initial_request"],
        "state_delta": {
          "initial_request": {
            "base_query_id": "single_0005",
            "hard_constraints": [
              "budget_constraint",
              "hotel_highest_rated",
              "restaurant_closest_to_attraction",
              "trip_date_range_required",
              "room_count_required",
              "intercity_round_trip_mode_required"
            ],
            "observable_profile_visible": true
          }
        },
        "must_update": ["initial_plan"],
        "response_expectation": "plan"
      },
      {
        "turn_id": 1,
        "role": "user",
        "utterance": "Please add \"Xianhu Botanical Garden\" to Day 3, while keeping every earlier requirement unchanged.",
        "sampled_deltas": ["add_db_grounded_attraction"],
        "state_delta": {
          "add_attraction": {
            "name": "Xianhu Botanical Garden",
            "insert_day": 3,
            "source": "city_context.signature_or_crowd_attractions"
          }
        },
        "must_preserve": [
          "budget_constraint",
          "hotel_highest_rated",
          "restaurant_closest_to_attraction",
          "trip_date_range_required",
          "room_count_required",
          "intercity_round_trip_mode_required"
        ],
        "must_update": ["add_attraction"],
        "resolution_need": "none",
        "response_expectation": "plan"
      }
    ],
    "active_constraints": {
      "route": {
        "origin_city": "Beijing",
        "destination_city": "Shenzhen",
        "intercity_mode": "flight",
        "round_trip_required": true
      },
      "dates": {
        "start_date": "2026-07-14",
        "end_date": "2026-07-17",
        "duration_days": 4,
        "return_date": "2026-07-17"
      },
      "party": {
        "travelers": 2,
        "rooms": 1
      },
      "budget_rmb": {
        "min": 5400,
        "max": 7200,
        "scope": "total estimated trip budget"
      },
      "lodging": {
        "city": "Shenzhen",
        "requirement": "highest-rated hotel in the destination city",
        "nights": 3
      },
      "meal": {
        "count": 1,
        "near": "Nantou Ancient City Museum",
        "selection_rule": "closest restaurant near the specified attraction"
      },
      "required_attractions": [
        {"name": "Nantou Ancient City", "source": "initial request"},
        {"name": "Xianhu Botanical Garden", "day": 3, "source": "turn 1 update"}
      ]
    },
    "preservation_contract": {
      "must_preserve": [
        "budget_constraint",
        "hotel_highest_rated",
        "restaurant_closest_to_attraction",
        "trip_date_range_required",
        "room_count_required",
        "intercity_round_trip_mode_required"
      ],
      "must_update": ["add_attraction"],
      "forbidden_shortcuts": ["drop_existing_hard_constraints"]
    }
  }
}
\end{lstlisting}
\end{tcolorbox}

\vspace{4mm}
\begin{tcolorbox}[
    promptbox={Plan Field Example},
    width=\linewidth,
    breakable,
    enhanced,
    boxsep=2pt,
    left=4pt,
    right=4pt,
    top=4pt,
    bottom=4pt,
    before skip=4pt,
    after skip=6pt
]
\footnotesize
\begin{lstlisting}[style=promptlisting]
"plan": """
<plan>
Day 1 (2026-07-14):
Current City: from Beijing to Shenzhen
Accommodation: Orange Hotel (Shenzhen Nanshan Xili Metro Station Branch), ¥354/room/night
08:00-09:30 | buffer | Arrive Beijing Capital International Airport, security and boarding
09:30-12:45 | travel_intercity_public | Flight HU7703, Beijing Capital International Airport - Bao'an International Airport, Economy, ¥1075/person
12:45-13:15 | buffer | Deplaning and baggage claim at Bao'an International Airport
13:15-13:45 | travel_city | Bao'an International Airport - Orange Hotel (Shenzhen Nanshan Xili Metro Station Branch), taxi, 14.85 km, 30 min, ¥36.1
13:45-14:15 | hotel | check-in, Orange Hotel (Shenzhen Nanshan Xili Metro Station Branch)
14:15-14:18 | travel_city | Orange Hotel (Shenzhen Nanshan Xili Metro Station Branch) - Dianbai Duck Porridge Shop, walking, 287 m, 3 min, ¥0
14:18-15:30 | meal | lunch, Dianbai Duck Porridge Shop, ¥61/person
15:30-15:33 | travel_city | Dianbai Duck Porridge Shop - Orange Hotel (Shenzhen Nanshan Xili Metro Station Branch), walking, 287 m, 3 min, ¥0
15:33-18:00 | hotel | rest, Orange Hotel (Shenzhen Nanshan Xili Metro Station Branch)
18:00-18:21 | travel_city | Orange Hotel (Shenzhen Nanshan Xili Metro Station Branch) - Nantou Ancient City, subway Line 5/Huanzhong - Line 12/Nanbao, 4.92 km, 21 min, ¥3/person
18:21-19:21 | attraction | Nantou Ancient City, ¥0/person
19:21-19:44 | travel_city | Nantou Ancient City - Hao Hao Wei Noodle House (Taoyuan Branch), walking, 1.91 km, 23 min, ¥0
19:44-21:00 | meal | dinner, Hao Hao Wei Noodle House (Taoyuan Branch), ¥26/person
21:00-21:25 | travel_city | Hao Hao Wei Noodle House (Taoyuan Branch) - Orange Hotel (Shenzhen Nanshan Xili Metro Station Branch), subway Line 12/Nanbao - Line 5/Huanzhong, 6.26 km, 25 min, ¥3/person

Day 2 (2026-07-15):
Current City: Shenzhen
Accommodation: Orange Hotel (Shenzhen Nanshan Xili Metro Station Branch), ¥354/room/night
10:00-11:14 | travel_city | Orange Hotel (Shenzhen Nanshan Xili Metro Station Branch) - Dameisha Coastal Boardwalk, taxi, 37.16 km, 74 min, ¥85.2
11:14-13:00 | attraction | Dameisha Coastal Boardwalk, ¥0/person
13:00-14:14 | travel_city | Dameisha Coastal Boardwalk - Dianbai Duck Porridge Shop, taxi, 36.88 km, 74 min, ¥84.5
14:14-15:30 | meal | lunch, Dianbai Duck Porridge Shop, ¥61/person
15:30-18:30 | hotel | rest, Orange Hotel (Shenzhen Nanshan Xili Metro Station Branch)
18:30-18:33 | travel_city | Orange Hotel (Shenzhen Nanshan Xili Metro Station Branch) - Dianbai Duck Porridge Shop, walking, 287 m, 3 min, ¥0
18:33-20:00 | meal | dinner, Dianbai Duck Porridge Shop, ¥61/person
20:00-20:03 | travel_city | Dianbai Duck Porridge Shop - Orange Hotel (Shenzhen Nanshan Xili Metro Station Branch), walking, 287 m, 3 min, ¥0

Day 3 (2026-07-16):
Current City: Shenzhen
Accommodation: Orange Hotel (Shenzhen Nanshan Xili Metro Station Branch), ¥354/room/night
09:30-10:17 | travel_city | Orange Hotel (Shenzhen Nanshan Xili Metro Station Branch) - Xianhu Botanical Garden, taxi, 23.71 km, 47 min, ¥55.6
10:17-12:17 | attraction | Xianhu Botanical Garden, ¥0/person
12:17-13:10 | travel_city | Xianhu Botanical Garden - Nantou Ancient City, taxi, 26.39 km, 53 min, ¥61.5
13:10-14:10 | attraction | Nantou Ancient City, ¥0/person
14:10-14:33 | travel_city | Nantou Ancient City - Hao Hao Wei Noodle House (Taoyuan Branch), walking, 1.91 km, 23 min, ¥0
14:33-15:45 | meal | lunch, Hao Hao Wei Noodle House (Taoyuan Branch), ¥26/person
15:45-16:10 | travel_city | Hao Hao Wei Noodle House (Taoyuan Branch) - Orange Hotel (Shenzhen Nanshan Xili Metro Station Branch), subway Line 12/Nanbao - Line 5/Huanzhong, 6.26 km, 25 min, ¥3/person
16:10-18:30 | hotel | rest, Orange Hotel (Shenzhen Nanshan Xili Metro Station Branch)
18:30-18:33 | travel_city | Orange Hotel (Shenzhen Nanshan Xili Metro Station Branch) - Dianbai Duck Porridge Shop, walking, 287 m, 3 min, ¥0
18:33-20:00 | meal | dinner, Dianbai Duck Porridge Shop, ¥61/person
20:00-20:03 | travel_city | Dianbai Duck Porridge Shop - Orange Hotel (Shenzhen Nanshan Xili Metro Station Branch), walking, 287 m, 3 min, ¥0

Day 4 (2026-07-17):
Current City: from Shenzhen to Beijing
Accommodation: -
10:00-10:30 | hotel | check-out, Orange Hotel (Shenzhen Nanshan Xili Metro Station Branch)
10:30-11:11 | travel_city | Orange Hotel (Shenzhen Nanshan Xili Metro Station Branch) - Bao'an International Airport, subway Line 5/Huanzhong - Line 1/Luo Bao, 15.41 km, 41 min, ¥5/person
11:11-15:30 | buffer | Airport check-in, security, and boarding wait at Bao'an International Airport
15:30-17:00 | buffer | Continue waiting at gate before boarding
17:00-20:30 | travel_intercity_public | Flight CA1342, Bao'an International Airport - Beijing Capital International Airport, Economy, ¥1200/person
20:30-21:00 | buffer | Deplaning and baggage claim at Beijing Capital International Airport

**Budget Summary**:
**Transportation: ¥4900.9**. Flights: (1075+1200) * 2 people = ¥4550. Intracity transport = ¥350.9.
**Accommodation: ¥1062**. ¥354 * 1 room * 3 nights.
**Meals: ¥592**. (61+26+61+61+26+61) * 2 people.
**Attractions & Tickets: ¥0**. Nantou Ancient City, Dameisha Coastal Boardwalk, and Xianhu Botanical Garden are free.
**Other: 0 RMB**
**Total Estimated Budget: ¥6554.9**
</plan>
"""
\end{lstlisting}
\end{tcolorbox}

\clearpage
\twocolumn

\clearpage
\newpage

\end{document}